\documentclass[10pt,twocolumn,letterpaper]{article}

\usepackage[pagenumbers]{cvpr} 

\usepackage[dvipsnames]{xcolor}

\usepackage[accsupp]{axessibility}
\usepackage{graphicx}
\usepackage{amsmath}
\usepackage{amssymb}
\usepackage{booktabs}
\usepackage{comment}
\usepackage{multirow}
\usepackage{float}

\definecolor{cvprblue}{rgb}{0.21,0.49,0.74}
\usepackage[pagebackref,breaklinks,colorlinks,citecolor=cvprblue]{hyperref}

\usepackage[capitalize]{cleveref}
\crefname{section}{Sec.}{Secs.}
\Crefname{section}{Section}{Sections}
\Crefname{table}{Table}{Tables}
\crefname{table}{Tab.}{Tabs.}
\crefname{figure}{Fig.}{Figs.}

\DeclareMathOperator*{\argmin}{argmin}

\def\confName{CVPR}
\def\confYear{2024}

\makeatletter
\def\thanks#1{\protected@xdef\@thanks{\@thanks
        \protect\footnotetext{#1}}}
\makeatother

\begin{document}

\title{From Variance to Veracity: Unbundling and Mitigating Gradient Variance in Differentiable Bundle Adjustment Layers}

\author{
    Swaminathan Gurumurthy\textsuperscript{1}, \ 
    Karnik Ram\textsuperscript{1 $\to$ 2}, \ 
    Bingqing Chen\textsuperscript{3},\\ 
    Zachary Manchester\textsuperscript{1}, \ \
    Zico Kolter\textsuperscript{1,3}%
    \\
    \textsuperscript{1}Carnegie Mellon University\ \
    \textsuperscript{2}TU Munich\\ 
    \textsuperscript{3}Bosch Center for Artificial Intelligence\\
    \thanks{Corresponding author: gauthamsindia95@gmail.com \\
    \indent \indent Code: \url{https://github.com/swami1995/V2V}}
    \vspace{-20pt}
}

\maketitle

\begin{abstract}
    Various pose estimation and tracking problems in robotics can be decomposed into a correspondence estimation problem (often computed using a deep network) followed by a weighted least squares optimization problem to solve for the poses. Recent work has shown that coupling the two problems by iteratively refining one conditioned on the other's output yields SOTA results across domains. However, training these models has proved challenging, requiring a litany of tricks to stabilize and speed up training. In this work, we take the visual odometry problem as an example and identify three plausible causes: (1) flow loss interference, (2) linearization errors in the bundle adjustment (BA) layer, and (3) dependence of weight gradients on the BA residual. We show how these issues result in noisy and higher variance gradients, potentially leading to a slow down in training and instabilities.
    We then propose a simple, yet effective solution to reduce the gradient variance by using the weights predicted by the network in the inner optimization loop to weight the correspondence objective in the training problem. This helps the training objective `focus' on the more important points, thereby reducing the variance and mitigating the influence of outliers. 
    We show that the resulting method leads to faster training and can be more flexibly trained in varying training setups without sacrificing performance. In particular we show $2$--$2.5\times$ training speedups over a baseline visual odometry model we modify.
\end{abstract}
\vspace{-20pt}
\section{Introduction}
\label{sec:intro}

Ego and exo pose estimation are essential for agents to safely interact with the physical world. These tasks have a long history of being tackled using geometry-based optimizaton \cite{mur2015orb, engel2014lsd, engel2017direct, lepetit2009ep}, and in the last decade, using deep networks to directly map inputs to poses \cite{deepvo, tartanvo, sfmlearner, pix2pose}. 
However, both these classes of approaches have shown brittleness ‒- not being robust to outliers in the data or having poor accuracy in unseen scenes.


\begin{figure}[t]
\centering
    \begin{subfigure}[t]{0.45\textwidth}
        \centering
        \includegraphics[width=\textwidth]{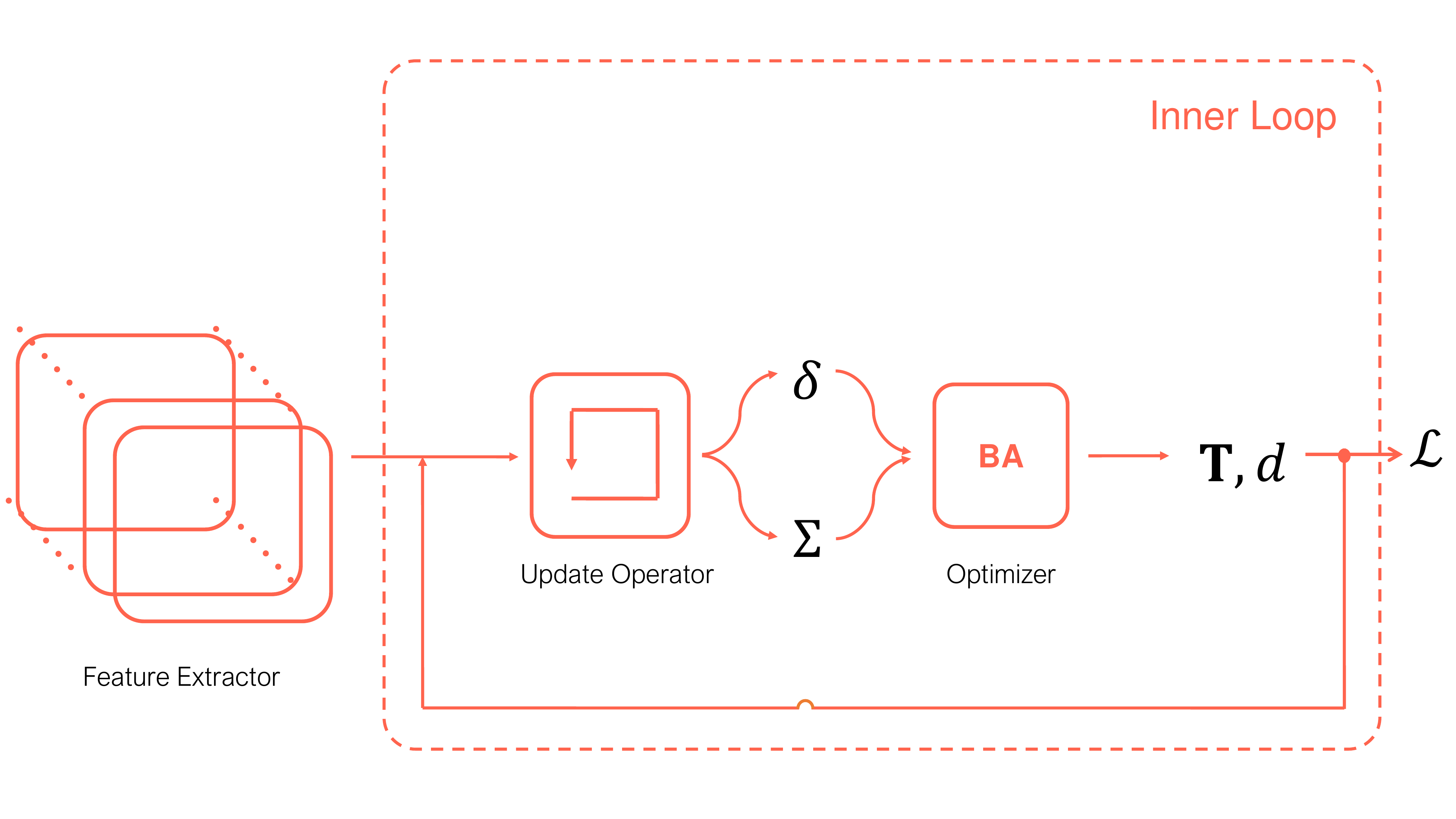}
        \caption{SOTA pose estimation methods~\cite{teed2021droid, teed2022deep, lipson2022coupled} tightly-couple learned front-ends with traditional BA optimizers. However, they can be slow to converge.}
        \label{fig:1a}
        \vspace{0.5em}
    \end{subfigure}
    \begin{subfigure}[t]{0.49\textwidth}
        \centering
        \includegraphics[width=1.0\textwidth]{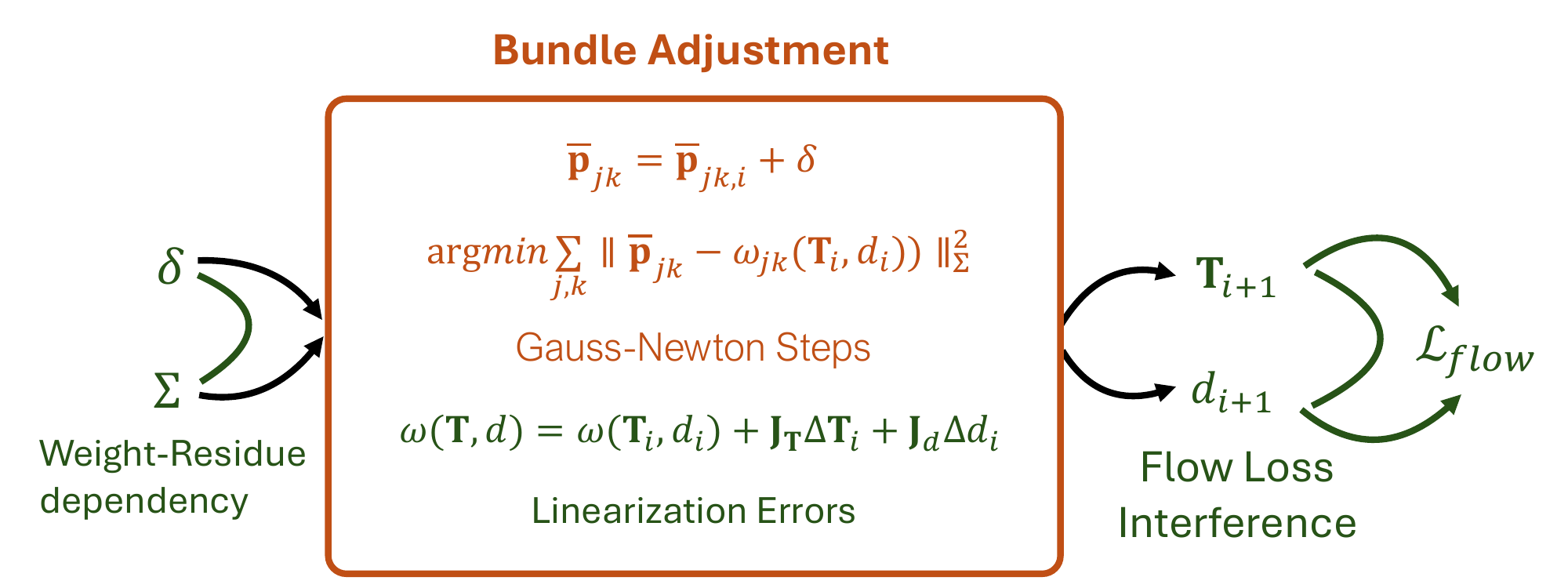}
        \caption{We identify three factors that lead to high variance in gradients during their training.}
        \label{fig:1b}
    \end{subfigure}
    \caption{We propose a simple, yet effective solution to stabilize and speed-up the training of SOTA pose estimation methods. (b) We first analyze the causes for their instability related to variance in their gradients, and (a) then mitigate them by using weights from the inner-loop  optimization to weigh the correspondence outer objective, which leads to improved performance.}
    \vspace{-1.5em}
\end{figure}

More recently, approaches that combine the best of both worlds in \textit{learning to optimize} have demonstrated substantially better performance than previous methods \cite{teed2021droid, teed2022deep, lipson2022coupled}.  These approaches combine a learned iterative update operator that mimics an optimization algorithm with implicit layers that enforce known geometric constraints on the outputs. This general architecture has appeared across many tasks even beyond pose estimation \cite{adler2017solving, adler2018learned, bhardwaj2020differentiable, fu2023batch}, where in each case an accurate and robust task-specific optimization solver is learned. In ~\cite{teed2021droid, teed2022deep, lipson2022coupled}, for the task of pose estimation, a recurrent network that iteratively updates pose and depth is learned through a differentiable weighted bundle adjustment (BA) layer that constrains the updates. Feature correspondences are also iteratively refined together with the poses, thereby dynamically removing outliers and leading to better accuracy.

Although these methods achieve state-of-the-art (SOTA) results, they take exceedingly long to train.  \cite{teed2021droid} mention that DROID-SLAM takes 1.5 weeks to train with 4x RTX 3090, while \cite{teed2022deep} mention that DPVO takes 3.5 days to train on a RTX 3090. Likewise, in our experiments, training the object pose estimation method from \cite{lipson2022coupled} took 1 week with 2x RTX 6000 for the smallest dataset reported in their paper. 

In this paper, we first investigate the reasons for the slow training convergence speeds of these methods, using \textit{deep patch visual odometry} (DPVO) \cite{teed2022deep} as an example problem setting for this analysis. We find that the bundle adjustment layer and the associated losses used in this setting lead to a high variance in the gradients. We identify three reasons contributing to the high variance. First, improper credit assignment arising from the specific choice of flow loss used which leads to interference between the gradients of outlier and inlier points. Second, improper credit assignment arising from the linearization issues in the bundle adjustment layer. And lastly, the dependence of the weight gradients on the residual of the BA objective resulting in the outliers dominating those gradients. We show how each of these problems lead to an increase in the gradient variance. 

Next, we leverage the analysis to propose a surprisingly simple solution to reduce the variance in gradients by weighting the flow loss according to the `importance' of the points for the problem, resulting in significant improvements in training speed and stability while achieving better pose estimation accuracy. We also experiment with other variance reduction techniques and demonstrate the superior performance of our proposed solution (Appendix \autoref{sec:var-abl}). 
Using DPVO as an example, we demonstrate 2-2.5x speedups with these simple modifications. Furthermore, we show that the modifications also make the training less sensitive to specific training setups. As a result, we are able to train in a non-streaming setting, while reaching similar accuracies in the streaming setting, thereby leading to a further 1.2-1.5x speedup in training. Lastly, we apply the modifications to DROID-SLAM \cite{teed2021droid} with little hyperparameter tuning to show that the proposed modifications transfer to a completely new pipeline providing similar speedups and stable training. Furthermore, we show that our best models achieve about $~50\%$ improvement on the TartanAir validation set and a $~24\%$ improvement on the TartanAir test set.
To summarize, the contributions of this paper are as follows:
\begin{itemize}
    \item We identify three candidate reasons for high variance in the gradients when differentiating through the BA problems for Visual Odometry (VO) and SLAM and show how they are all affected by the presence of outliers.
    \item Using DPVO \cite{teed2022deep} as an example VO pipeline, we propose a simple modification to the loss function that reduces the variance in the gradients by mitigating the effect of outliers on the objective.
    \item We show that the above modification results in significant speedups and improvements in accuracy of the model on the TartanAir~\cite{wang2020tartanair} validation and test splits used in the CVPR 2020 challenge. Further, we show that the modifications can be applied out-of-the-box to other settings/methods that use differentiable BA layers, such as DROID-SLAM and the non-streaming version of DPVO to obtain similar benefits.
\end{itemize}

\section{Related Work} 
\paragraph{Pose Estimation using Deep Learning.} A large body of works have tackled pose estimation, and we describe a few representative works that use deep learning here. For a broader overview, we refer the reader to \cite{cadena2016past, chen2020survey, fan2022deep}. \cite{deepvo, sfmlearner, tartanvo} proposed deep networks to directly estimate ego pose between pairs of frames. \cite{sarlin2020superglue, min2020voldor, koestler2022tandem, tateno2017cnn} integrate learned representations (features or depth) into traditional ego-pose estimation pipelines. \cite{tang2018ba, deepfactors, d3vo} imposed geometric constraints on ego-pose network outputs via differentiable optimization layers. Similar approaches have been proposed for the task of multi-object pose estimation where 2D-3D correspondences are directly regressed~\cite{bb8, pix2pose} and then passed through a differentiable PnP solver~\cite{bpnp} for pose estimation. Overall, these works showed that deep learning could be applied to these tasks but fell short in accuracy and generalization.

Optimization-inspired iterative refinement methods have been applied to ego-pose~\cite{zhou2018deeptam, Ummenhofer_2017_CVPR, clark2018learning, jatavallabhula2020slam} and exo-pose estimation~\cite{iwase2021repose, labbe2020cosypose} where the network iteratively refines its pose estimates as an update operator in order to satisfy geometric constraints. More recently, methods that iteratively refine poses and correspondences in a tightly-coupled manner have been proposed~\cite{teed2021droid, teed2022deep, lipson2022coupled}. In these works, a network predicts patch correspondences~\cite{teed2022deep} or dense flow~\cite{teed2021droid, lipson2022coupled} which are then updated together with poses and depths in an alternating manner where one feeds into the other through differentiable geometric operations. In addition to correspondences, these methods also predict weights for the correspondences which have been shown to be important for pose estimation accuracy in many independent works~\cite{kanazawa,ranftl2018deep,burnett_rss21,muhle2023learning}. Overall, these iterative methods have achieved impressive performance in terms of accuracy and generalization, but they still need large GPU memories \cite{teed2022deep} and their training times are prohibitively long which has limited their adoption for research.

\vspace{-15pt}
\paragraph{Challenges with Implicit Optimization Layers.} With the advent of implicit layers, it is possible to incorporate an optimization problem as a differentiable layer \cite{amos2017optnet, agrawal2019differentiable, pineda2022theseus}, which captures complex behaviours in a neural network. The BA layer~\cite{tang2018ba} used in this work is an instance of such layers. In the forward pass, an implicit optimization layer solves a regular optimization problem given the current estimate of problem parameters. In the backward pass, one differentiates through the KKT conditions of the optimization problem to update the problem parameters. 

While these implicit optimization layers boast expressive representational power, there exist challenges with such layers. Firstly, these problems naturally take on a bilevel structure, where the inner optimization learns the problem parameters and the outer problem optimizes for the decision variables given the current estimation of problem parameters. As a result, these problems are inherently hard to solve\cite{amos2018differentiable,amos2017optnet,howell2022dojo}, as their easiest instantiation, e.g., linear programs for both inner and outer problems, can be non-convex \cite{beck2021gentle}. While the convergence issues may be alleviated by techniques such as using good initialization \cite{amos2017optnet} or robust solvers, there does not exist a general solution to the authors' best knowledge. Secondly, a range of numerical issues can arise from implicit optimization layers. The gradients derived from KKT conditions are only valid at fixed points of the problem. In practice, the solver may need to run long enough to reach a fixed point or a fixed point may not exist at all \cite{donti2021dc, amos2018differentiable}. The problem may be ill-conditioned due to reasons such as stiffness or discontinuities from physical systems \cite{suh2022differentiable} or compounding of gradients in chaotic systems \cite{metz2021gradients}. A number of problem-specific solutions have been proposed \cite{suh2022differentiable, howell2022calipso, howell2022dojo} to these problems. For example, \cite{suh2022differentiable, antonova2023rethinking} use zeroth-order methods to deal with non-smoothness and non-convexity in the problem. \cite{howell2022calipso, howell2022dojo} use interior point relaxations to smooth the discontinuities. Similarly, \cite{donti2021dc, bianchini2023simultaneous} use penalty-based relaxations to handle the discontinuities. It's also common to regularize the inner problem during the backward pass to deal with ill-conditioning \cite{amos2018differentiable, fung2022jfb}. However, given the vastness of the problems, we are of the opinion that this is still a broadly under-studied area. 
\vspace{-5pt}
\section{Background}
\vspace{-5pt}
In this section, we review the approach of DPVO~\cite{teed2022deep} for iterative ego-pose estimation, which serves as an example setting for all our analysis and experiments.
\vspace{-10pt}
\paragraph{Feature Extraction.} A scene, as observed from an input video, is represented as a set of camera poses  $\mathbf{T}_j\in \mathbb{SE}(3)$ and square image patches $\mathbf{P}_k$. Patches are created by randomly sampling $2$D locations in the image and extracting $p \times p$ feature maps centered at these coordinates $\mathbf{p}_k$.
A bipartite patch-frame graph is constructed by placing an edge between every patch $k$ and each frame $j$ within distance $r$ of the patch source frame. The reprojections of a patch in all of its connected frames form the trajectory of the patch.
\vspace{-10pt}
\paragraph{Update Operator.}The update operator iteratively updates the optical flow of each patch over its trajectory. The operator updates the embedding of each edge $(k,j)$ of the patch graph via temporal convolutions and message passing. These updated embeddings are used by two MLPs to predict flow revisions $\delta_{jk} \in \mathbb{R}^2$ and confidence weights for each patch $\Sigma_{jk} \in \mathbb{R}^2$ between $[0,1]$. The flow revisions are used to update the reprojected patch coordinates $\bar{\mathbf{p}}_{jk}:=\bar{\mathbf{p}}_{jk} + \delta_{jk}$, which are passed to a differentiable BA layer along with their confidence weights $\Sigma_{jk}$.

\vspace{-10pt}
\paragraph{Differentiable Bundle Adjustment.} The bundle adjustment (BA) layer solves for the updated poses and depths that are geometrically consistent with the predicted flow revisions. The BA layer operates on a window of the patch graph to update the camera poses and patch depths, while keeping the revised patch coordinates $\bar{\mathbf{p}}_{jk}$ fixed. The BA objective is as follows: 

\begin{equation}\label{eqn:bundle_adjustment}
    \min_{\mathbf{T}_{ij}, d_k} \sum_{(k, j)} ||\bar{\mathbf{p}}_{jk}-\Pi(\mathbf{T}_{ij}, \Pi^{-1}(\mathbf{p}_k,d_{k}))||^2_{\Sigma_{jk}}
\end{equation}

where $\Pi$ denotes the projection operation, $d_k$ denotes the depth of the $k\textsuperscript{th}$ patch in the source frame $i$, and $\mathbf{T}_{ij}$ is the relative pose $\mathbf{T}_i\mathbf{T}_j^{-1}$. This objective is optimized using two Gauss-Newton iterations. The optimized poses and depths are then passed back to the update operator to revise the patch coordinates, and so on in an alternating manner.
\vspace{-10pt}
\paragraph{Training Loss.}
The network is supervised using a flow loss and pose loss computed on the intermediate outputs of the BA layer. The flow loss computes the distance between the ground truth patch coordinates and estimated patch coordinates over all the patches and frames:
\begin{equation}\label{eq:floss}
    \mathcal{L}_{\text{flow}} = \sum_{j,k} \Vert \mathbf{p}^*_{jk} - \mathbf{\hat{p}}_{jk}\Vert_2
\end{equation}    
where $\mathbf{\hat{p}}_{jk} = \Pi(\mathbf{T}_{ij}, \Pi^{-1}(\mathbf{p}_k,d_{k}))$ and $\mathbf{p}^*_{jk}$ is the corresponding reprojection of patch $k$ in frame $j$ using the ground truth pose and depth. Note that this loss amounts to a difference in the patch coordinates and not in the flows as the source patch coordinates in each flow term cancel out.

The pose loss is the error between the ground truth poses $\mathbf{G}$ and estimated poses $\mathbf{T}$ for every pair of frames $(i,j)$:

 \begin{equation}
    \mathcal{L}_{\text{pose}} = \sum_{(i,j)} \Vert \text{Log}_{\mathbb{SE}(3)}[(\mathbf{G}_i.\mathbf{G}_j^{-1})^{-1}.(\mathbf{T}_i.\mathbf{T}_j^{-1})] \Vert_2
 \end{equation}

The total loss is a weighted combination of the flow loss and pose loss,
\begin{equation}\label{eq:dpvoloss}
\mathcal{L} = 10\mathcal{L}_{\text{flow}} + 0.1\mathcal{L}_{\text{pose}}
\end{equation}

The original DPVO model is trained on random sequences of $15$ frames, where the first $8$ frames are used together for initialization and the subsequent frames are added one at a time. Their model is trained for $240$K iterations using $19$GB of GPU memory which takes $3.5$ days on an RTX $3090$. A total of $18$ iterations of the update operator is applied on each sequence, where the first $8$ iterations are applied during initialization as a batch-optimization, and the subsequent iterations are for every new, added frame. In our paper, we refer to these update iterations as the `inner-loop optimization', this mode of training as the `streaming' setting, and training models in our experiments to only batch-optimize the first $8$ frames as the `non-streaming' setting.

\vspace{-5pt}
\section{Factors Affecting Training Convergence}\label{sec:factors}
In this section, we identify three possible causes for slow training convergence. We show how each of these result in noisier/higher variance gradients during training, and consequently result in instabilities and slowdowns.

\subsection{Flow loss interference}\label{sec:flowinterf}
The flow loss defined in \autoref{eq:floss} operates on the reprojected patch coordinates $\hat{\mathbf{p}}_{jk}$ which are computed using the optimized poses $\mathbf{T}_{i}, \mathbf{T}_j$ and depth $d_k$ outputs from the BA layer. Thus, the gradient of the loss with respect to $d_k$ (and similarly for poses $\mathbf{T}_i, \mathbf{T}_j$) can be written as follows,
\begin{subequations}
\begin{align} \label{eq:graddk}
\nabla_{d_k} \mathcal{L}_{\text{flow}} \propto \sum_j \nabla_{d_k} \Pi(\mathbf{T}_{ij}, (\mathbf{p}_k,d_k)).\\
\nabla_{T_i} \mathcal{L}_{\text{flow}} \propto \sum_{k,j} \nabla_{T_i} \Pi(\mathbf{T}_{ij}, (\mathbf{p}_k,d_k)).
\end{align}
\end{subequations}
Thus, the gradients with respect to each reprojected patch $\hat{\mathbf{p}}_{jk}$ gets aggregated in the computation graph at the corresponding depth $d_k$ (likewise for poses $\mathbf{T}_i, \mathbf{T}_j$) at the output of the BA layer. This becomes problematic when a significant fraction of the projections are noisy/outliers, 
as the noisy/outlier gradients would dominate the inlier gradients in the sum in \autoref{eq:graddk}, leading to more noise in the total gradient estimate. 

Since these gradients are also backpropagated through the BA layer, it results in noisy gradient estimates for the network parameters as well. Specifically, in the BA layer, each $d_i$/$\mathbf{T}_i$ are again a function of all the predicted flows and weights associated with that point/frame. Thus, the same noisy gradient computed at $d_i$/$\mathbf{T}_i$, gets backpropagated to all the associated points. This leads to the gradient estimates being noisy even at the `good' predictions by the network.

\subsection{Linearization errors in BA gradient} \label{sec:lingrads}
Given gradient estimates at the output poses and depth of the BA layer $\nabla_{d}\mathcal{L}, \nabla_{\mathbf{T}}\mathcal{L}$, the gradients with respect to its input flows and weights are computed as follows:
\begin{subequations}
\begin{align}\label{eq:flgrads}
    \nabla_\delta \mathcal{L} &= -(\nabla_{\mathbf{T}}\mathcal{L})^T(\mathbf{J}_{\mathbf{T}}^T\Sigma \mathbf{J}_{\mathbf{T}})^{-1}\mathbf{J}_{\mathbf{T}}^T\Sigma \nonumber
    \\&
    \quad -(\nabla_{d}\mathcal{L})^T(\mathbf{J}_{d}^T\Sigma \mathbf{J}_{d})^{-1}\mathbf{J}_{d}^T\Sigma
\end{align}
\begin{align}\label{eq:wt_grads}
    \nabla_\Sigma \mathcal{L} &= -(\nabla_{\mathbf{T}}\mathcal{L})^T(\mathbf{J}_{\mathbf{T}}^T\Sigma \mathbf{J}_{\mathbf{T}})^{-1}\mathbf{J}_{\mathbf{T}}^T \text{diag}(\mathbf{r})\nonumber
    \\&
    \quad -(\nabla_{d}\mathcal{L})^T(\mathbf{J}_{d}^T\Sigma \mathbf{J}_{d})^{-1}\mathbf{J}_{d}^T\text{diag}(\mathbf{r}) 
\end{align}
\end{subequations}

where, $\mathbf{r}=(\bar{\mathbf{p}}_{kj}-\hat{\mathbf{p}}_{kj})$ is the bundle adjustment residual, $\mathbf{J}_d$ and $\mathbf{J}_{\mathbf{T}}$ are the jacobians of the projection $\Pi(\mathbf{T}_{ij}, \Pi^{-1}(\mathbf{p}_k,d_{k}))$ with respect to depth $d$ and pose $\mathbf{T}$ respectively. This expression can be derived by applying the implicit function theorem (Theorem 1B.1) \cite{dontchev2009implicit},   on the BA problem as shown in Appendix \autoref{sec:ift}.

Since the projection is non-linear containing multiple multiplicative operations, we observe that the Jacobians $\mathbf{J}_d$ and $\mathbf{J}_{\mathbf{T}}$ themselves are a function of $d$ and $\mathbf{T}$. Thus, a high variance in the initialized $d$ or $\mathbf{T}$ naturally lead to a high variance in the Jacobians, thereby leading to a high variance in the corresponding gradients $\nabla_{\Sigma}$ and $\nabla_{\delta}$, which are then backpropagated through the network. In our setup, $d$ is initialized to random values and $\mathbf{T}$ is initialized to identity. Thus, the variance from linearization is primarily contributed by the linearization around the current $d$. 

The use of a weighted objective in the BA problem partially mitigates this issue by masking out the gradients on the flows corresponding to the outlier points (which contribute the most to this high variance). However, the high variance remains problematic especially in the initial iterations of training (when the weight estimates themselves are not very accurate) and in the initial iterations of the inner-loop optimization when a large fraction of the depth and pose estimates are inaccurate.



\subsection{Dependence of weight gradients on the BA residual}\label{sec:wresidual}

In the previous section, we discussed the effect of outliers on the BA linearization and consequently on the gradients. However, outliers in the BA problem contribute to an increase in gradient variance in a more straightforward way.
Specifically, they have a direct effect on the gradient of the weights, as can be seen from the expression in \autoref{eq:wt_grads}.
The expression shows the direct dependence of the weight gradients on the residual, $\mathbf{r}=(\bar{\mathbf{p}}_{jk}-\hat{\mathbf{p}}_{jk})$, of the BA problem. Thus, the presence of high residual points in the optimization problem result in high variance in the weight gradients.

In fact, the presence of outliers also biases the weight gradients towards highly positive values as the training objective tries to reduce the influence of the outliers. This consequently leads to a collapse in the weight distribution. However, we observe that a straightforward fix used by prior work\cite{teed2021droid, teed2022deep}, i.e, clipping the magnitude of gradient passing through the weights easily mitigates this bias. We discuss more details on this effect with a simple illustrative example in Appendix \autoref{sec:wt-res}.



To summarize, the above section highlights various aspects of the existing setup that contribute to noisy/high variance gradients. The noise and high variance in gradient estimates leads to ineffective parameter updates, thereby leading to training instabilities and slowdown. Furthermore, it's also important to note that the aforementioned effects exacerbate each other. For example, worse weight estimates result in bad BA outputs, which in turn contribute to worsening the flow loss interference and BA linearization errors, which further leads to noisier gradients thereby slowing down weight/flow updates, thus repeating the vicious cycle. By the same argument, mitigating either of these effects can also provide significant improvements on other problems!
\vspace{-5pt}
\section{A very simple solution: Weighted flow loss}\label{sec:wtd-loss}
We start with observing that all three problems mentioned in the previous section get exacerbated by the presence of outliers or computing gradients through outliers. So the natural question is if there exist obvious solutions to mask out the outliers in the outer training problem. 

One of the tricks used by \cite{teed2021droid, teed2022deep} already partially accounts for this in the pose loss, i.e, they do not include the pose loss for the first couple of inner-loop iterations, thereby mitigating some of the issues discussed in \autoref{sec:lingrads}. This simple modification in \cite{teed2021droid, teed2022deep} seems to provide a significant boost in training speeds as we show in our ablation experiments in Appendix \autoref{sec:inner-pose}. 

Similar heuristics for the flow loss are harder to find as the depth/flow estimates of a significant fraction of points are bad even at the latter inner-loop iterations. Conventionally, SLAM and visual odometry problems define heuristic kernels on the flow residuals \cite{Barron_2019_CVPR, chebrolu2021adaptive} depending on the expected distribution of residuals/errors to trade-off between robustness and accuracy. Unfortunately, coming up with a similar simple/consistent heuristic to define `outliers' in the outer training problem is more challenging as the errors and distribution of errors vary across examples, training iterations and inner optimization iterations. This requires a heuristic that adapts to the specific example, training convergence, and inner-loop optimization iteration. 

Conveniently, we find that the weights learnt by the inner update operator for the bundle adjustment problem satisfy all these properties as they adapt online with the changing distribution of errors/residuals. Moreover, empirically we observe that the network learns a reasonable weight distribution very early on in training, while adapting the weight distribution rapidly to any changes in flows. Thus, we observe that  using these weights to weight the flow loss works surprisingly well. The resulting flow loss is as follows.
\begin{equation}\label{eq:wtdfloss}
    \mathcal{L}_{\text{flow}} = \sum_{j,k} \Vert \mathbf{p}^*_{jk} - \mathbf{\hat{p}}_{jk}\Vert_{\Sigma_{jk}^\perp}
\end{equation}    
where $\perp$ denotes the stop gradient operator to prevent the objective from directly driving the weights to zero (We provide more discussion on what factors prevent these weights from collapsing to zero in Appendix \autoref{sec:wt-collapse}). The main difference between this and \autoref{eq:floss} is that each residual in this objective is weighted by the weights $\Sigma_{jk}$ predicted by the network for the inner BA problem. Intuitively, this objective incentivizes the network to focus on the points which are important for the inner optimization problem at that optimizer step / training iteration for that particular example.

Although the modification seems trivial and obvious in hindsight, we observe that it is significantly more effective than various other (more complicated) variance reduction approaches we tried (studied in Appendix \autoref{sec:var-abl}). This apparent simplicity and effectiveness underscore the value of the proposed modifications!
\vspace{-15pt}
\paragraph{Balancing loss gradients.}
The introduction of the weighted flow loss changes the gradient contribution from the flow loss throughout training as the weight distribution changes. Thus, instead of using fixed coefficients to trade-off between pose and flow loss as in \autoref{eq:dpvoloss}, we periodically (every 50 training iterations) update the flow loss coefficient $\beta$ to ensure the gradient contributions of the pose and flow loss remain roughly equal throughout training. Given the infrequency in these updates, they barely affect the training speed and hence are cheap to compute amortized over the entire training run.

\begin{equation}
    \beta = \frac{\|\nabla_\theta \mathcal{L}_{\text{pose}}\|_2}{\|\nabla_\theta \mathcal{L}_{\text{flow}}\|_2}
\end{equation}
\begin{equation}
    \mathcal{L} = \mathcal{L}_{\text{pose}} + \beta \mathcal{L}_{\text{flow}}
\end{equation}

\begin{figure*}[t]
    \begin{subfigure}{0.31\linewidth}
        \centering
        \includegraphics[width=\linewidth]{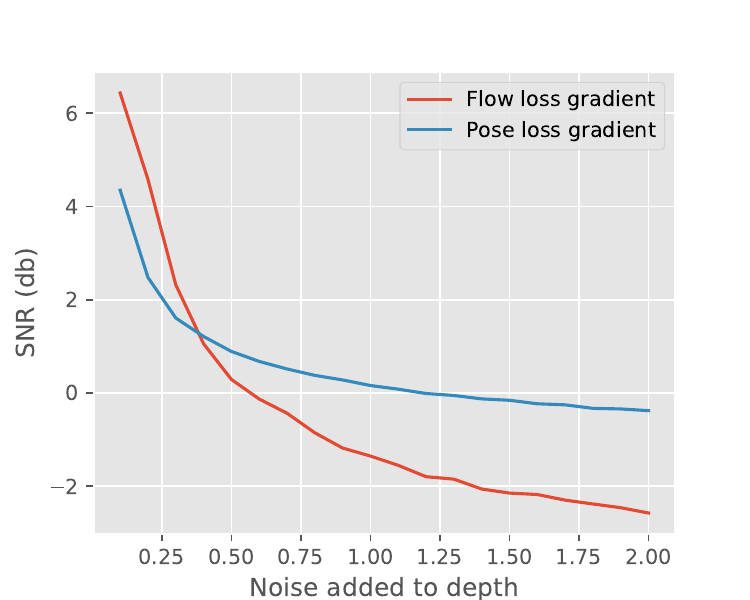}
        \caption{Impact of depth noise on linearization.}
        \label{fig:linearization}
    \end{subfigure}%
    \begin{subfigure}{0.31\linewidth}
        \centering
        \includegraphics[width=\linewidth]{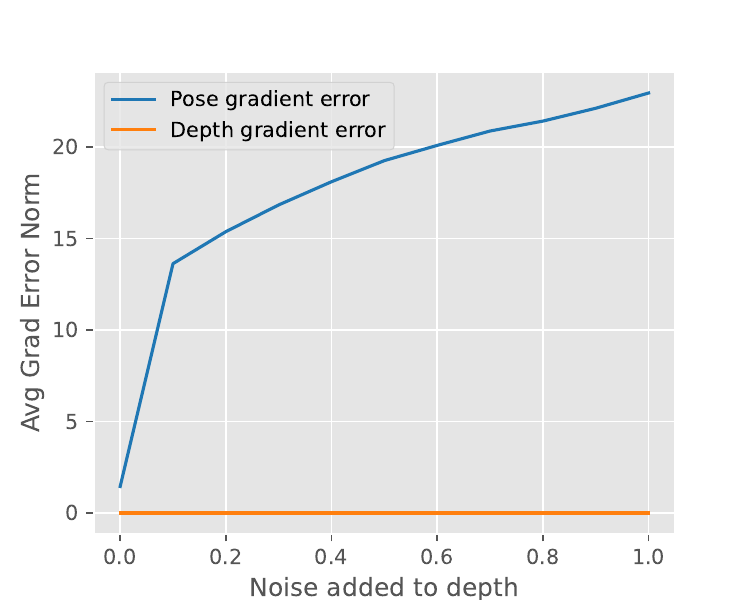}
        \caption{Impact of depth noise on flow loss gradients.}
        \label{fig:flowinterf}
    \end{subfigure}%
        \begin{subfigure}{0.37\linewidth}
        \centering
        \includegraphics[width=\linewidth]{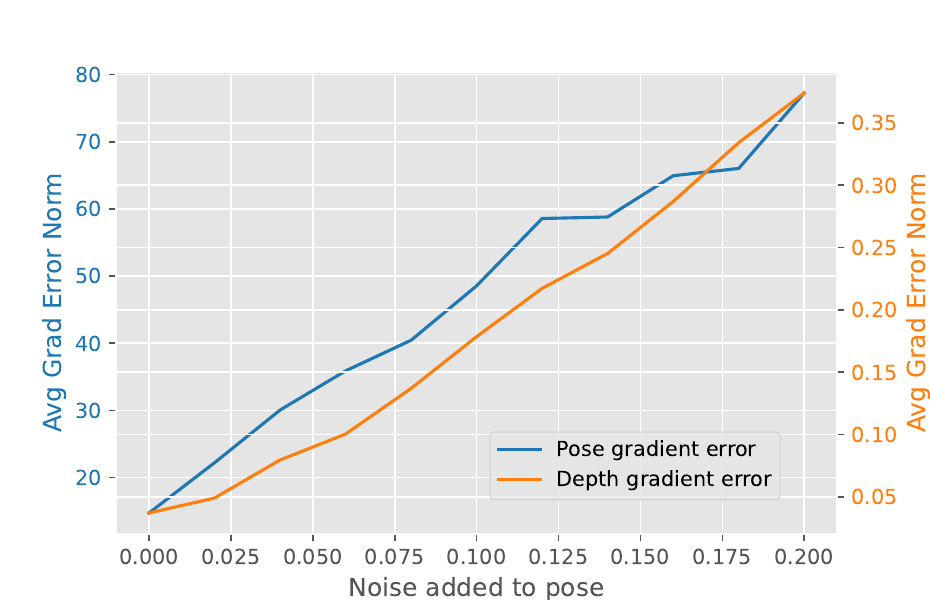}
        \caption{Impact of pose noise on flow loss gradients.}
        \label{fig:flowinterf2}
    \end{subfigure}
    \caption{(a) We compute the signal-to-noise ratio (SNR) in the loss gradients as we artificially add depth noise while linearizing the BA problem for gradient computation. We observe that the SNR in the flow loss deteriorates rapidly indicating its sensitivity to linearization errors. (b) We artificially add noise to a subset of depths right before the flow loss computation. We show the average gradient errors on all the pose and `clean' depth variables as a result of the added noise. We see a monotonic increase in gradient error in pose gradients as we increase the noise added showing the impact `outliers' have on the gradients of even the `inlier' variables. (c) Similar to (b), here we add noise to the the first frame's pose and show the gradient errors on the rest of the frames and depths.} 
    \vspace{-10pt}
\end{figure*}
\vspace{-10pt}
\section{Results and Analysis}

We analyze the effect of the factors discussed in ~\autoref{sec:factors} on the original DPVO model on the TartanAir~\cite{wang2020tartanair} dataset. We then analyze a version trained with our proposed weighted flow objective. We show that the weighted objective helps increase the signal to noise ratio in the gradients throughout training and show the improvements in performance as a result. We also evaluate the pose estimation performance of this version on the TartanAir~\cite{wang2020tartanair}, EuRoC~\cite{burri2016euroc}, and TUM-RGBD~\cite{sturm2012benchmark} benchmarks. We use the average absolute trajectory error (ATE) after $\text{Sim}(3)$ alignment of the trajectories, as the evaluation metric for pose estimation.

\subsection{Analyzing factors affecting gradient variance}
\vspace{-5pt}
To understand the impact of linearization on the gradient variance (\autoref{sec:lingrads}), we analyze the impact of adding noise to the depth used to compute the Jacobians in the BA problem. We leave the rest of the forward and backward pass unaltered and only add noise to the depth while computing the linearization for the backward pass in the BA problem. This helps us isolate the effects of linearization on the gradient computations. Specifically, \autoref{fig:linearization} shows the signal-to-noise ratio (SNR) of the flow and pose loss gradients with respect to $\delta$ with increasing levels of noise. The SNR is computed assuming the no-depth-noise gradient as the true signal and treating any deviations from it as noise. The SNR computation details are provided in Appendix \autoref{sec:snr}. This yields two interesting observations. First, the SNR deteriorates rapidly in the beginning indicating that the gradients are indeed sensitive to the noise in the iterates used for linearization. Second, the SNR in the flow loss gradients is high initially, but deteriorates rapidly compared to the pose loss gradients with increasing noise. This highlights the need to make flow loss robust to noisy points. 



To analyze the effect of flow loss on the gradient noise (\autoref{sec:flowinterf}), we introduce noise on a few depth points or a single frame pose right before computing the flow loss and study the effect of the noise on the gradients of all the other points/poses. \autoref{fig:flowinterf2} shows a monotonic increase in gradient errors on the depths as well as all poses as we increase the noise added to the first pose. Likewise, \autoref{fig:flowinterf} shows the monotonic increase in gradient errors of all poses as we add increasing amounts of noise to all depths on the first frame. This shows how outliers with increasingly large errors can have an increasingly adverse effect on the gradients of the non-outlier points/frames as well. The gradient errors are computed as the average L2 norm of the deviation in gradient from the no-noise gradients.

    

We analyze the weight residual dependence (\autoref{sec:wresidual}) and the resulting variance / bias in Appendix \autoref{sec:wt-res}, as its connections to the use of weighted loss are less direct. 

\vspace{-5pt}
\subsection{Effect of the weighted flow loss on training}
\begin{figure}
    \vspace{-10pt}
    \centering
    \includegraphics[scale=0.6]{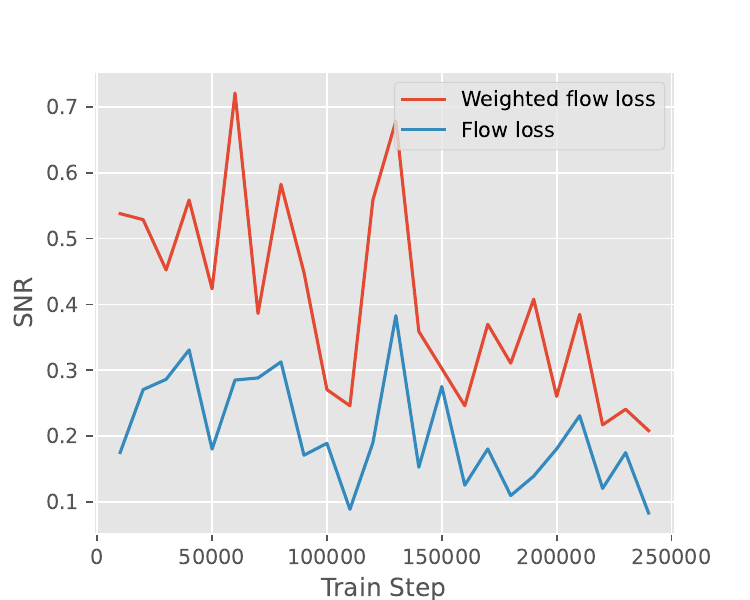}
    \caption{We compute the signal-to-noise ratio in the gradients of the flow loss and the weighted flow loss w.r.t flow network parameters at different training iterations of the base model. Specifically, we use the last linear layer's weights of the flow computation head of the network. We find that the weighted flow loss gradients have a higher SNR throughout the training. This is especially true in the initial iterations of training when the outlier count is very high. }
    \label{fig:snr-wtd}
    \vspace{-15pt}
\end{figure}

\vspace{-5pt}
To understand the effect of the weighted flow loss on the variance of the gradients, we study the SNR of the gradients on the flow network parameters. \autoref{fig:snr-wtd} shows the SNR with the flow loss and the weighted flow loss at different points during training. The SNR computation details are provided in Appendix \autoref{sec:snr}. The plots clearly demonstrate that the usage of weighted flow loss results in a boost in SNR throughout training. The boost is especially prominent in the initial stages of training, when the impact of outliers and noise in the pose/depth estimates are most significant. This clearly shows the promise of using the weighted flow loss instead of the regular flow loss for training.

\begin{figure}
    \centering
    \includegraphics[scale=0.6]{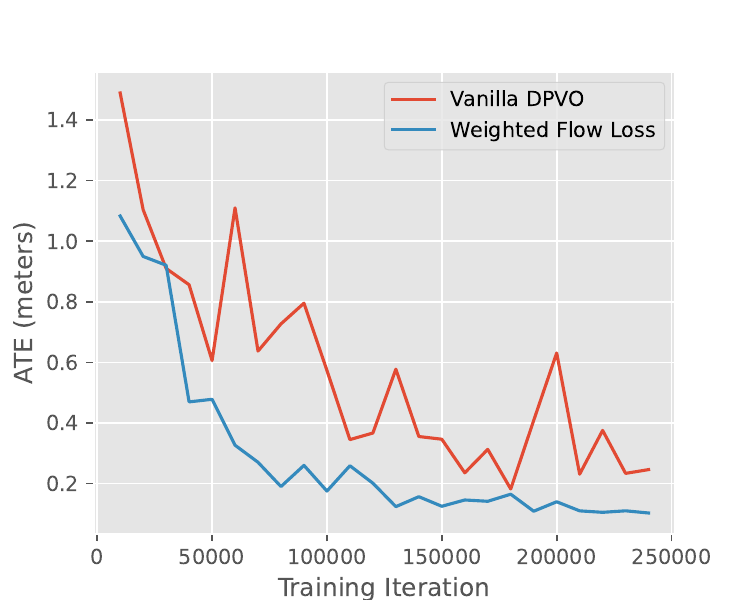}
    \caption{We observe that DPVO when trained with our weighted flow loss achieves much faster training, reaching $\sim\!0.2$ m accuracy in only $80$K iterations, and is much more stable. We report the median ATE across three trials on the validation split of TartanAir.}
    \label{fig:str-val-avg}
\end{figure}

\begin{figure}
    \vspace{-15pt}
    \centering
    \includegraphics[scale=0.6]{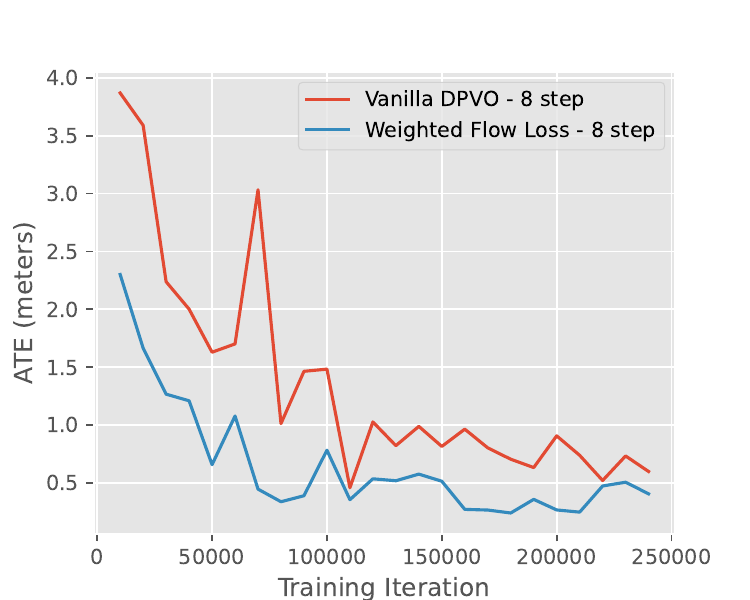}
    \caption{We retrain DPVO with and without the modified flow loss in the non-streaming batch setting and evaluate both models on validation sequences from TartanAir in the streaming setting. We observe that, beyond training faster and being more stable, the modified version generalizes better than the original model. This allows the model to be trained on shorter sequences without suffering high performance drops, thanks to the reduced gradient variance. We report the median ATE across three trials.}
    \label{fig:gl-val-avg}
    \vspace{-15pt}
\end{figure}

We retrain DPVO with our modified weighted flow loss on the TartanAir dataset and show its validation error performance across training iterations. We observe in \autoref{fig:str-val-avg} that the average ATE of our method drops rapidly compared to the original. While our model takes only $80$K iterations to reach an average error of $\sim\!0.2$ m, the original model reaches the same performance at $180$K iterations. In fact, while that's the peak performance reached by the base model, our model continues to improve and reaches a final convergence error of $\sim\!0.10$ m, achieving half the base model's convergence error on the validation set. We also observe that, unlike the original model, the errors don't fluctuate rapidly over epochs and is more stable.

Further, the reduced variance in gradients allows us to train in other  setups as well. For example, \autoref{fig:gl-val-avg} shows the ATE of our model against the base model trained in the non-streaming setting, i.e. using just $8$ frame initialization sequences instead of $15$ frame sequences. This allows the models to be trained faster (with per iteration cost of $0.6$s vs $1.6$s for the streaming version on an RTX A6000 GPU) and with lower GPU memory ($7.2$GB as opposed to $19.2$GB GPU memory). Note that the evaluations are still done as earlier, i.e, by rolling out the model on the full validation sequences in the streaming setting. Yet, we observe that despite being trained to only batch-optimize over $8$ frames, it generalizes to the streaming setting with our modified models obtaining a peak performance of $\sim\!0.2$m pose errors in $180$K iterations (i.e, $2.7\times$ faster than the base streaming model). Furthermore, we also test our modifications on DROID-SLAM, a completely different pipeline that also uses Bundle Adjustment layers with little to no hyperparameter tuning. We present the results in Appendix \autoref{sec:droid}. Again, we observe that the modifications result in significant speedups and stability during training suggesting that the methods and analysis discussed in this paper applicable broadly to approaches using differentiable BA layers.





\begin{figure}
    \vspace{-3pt}
    \centering
    \includegraphics[scale=0.6]{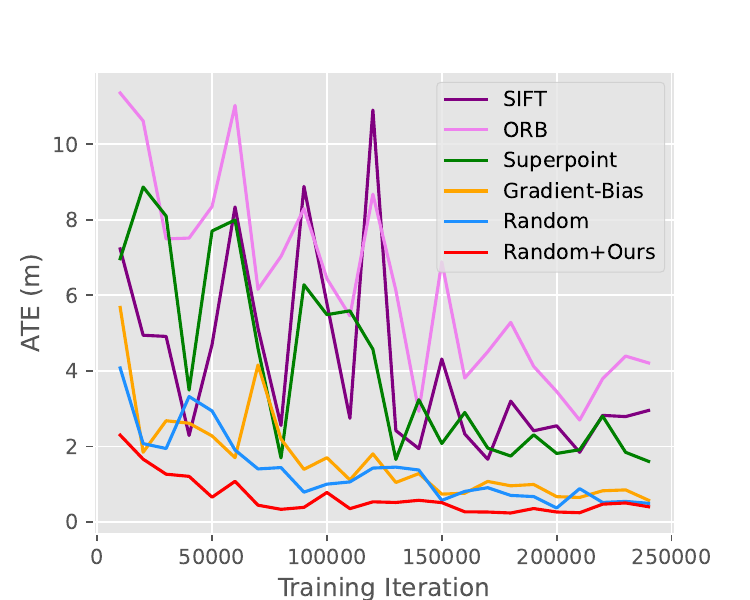}
    \vspace{-8pt}
    \caption{We retrain the original DPVO model with standard feature detection methods and observe that our method of random sampling with the modified flow objective has much improved training convergence. We report the median ATE across three trials on the validation split of TartanAir.}
    \label{fig:features}
    \vspace{-15pt}
\end{figure}

Finally, to evaluate the ability of our modified model to weight patches effectively, we compare against other standard methods for selecting patches. Specifically, we re-train the original DPVO model with patches selected using SIFT~\cite{lowe2004distinctive}, ORB\cite{rublee2011orb}, Superpoint\cite{detone2018superpoint}, and naive gradient based sampling instead of the default random sampling. As shown in ~\autoref{fig:features}, we observe that random sampling along with the weights learned by our network is much more stable and accurate than other patch selection methods.

\begin{table*}[!h]
\centering
\resizebox{0.9\linewidth}{!}{%
\begin{tabular}{lcccccccc|cccccccc|c}
\toprule
& ME & ME & ME & ME & ME & ME & ME & ME & MH & MH & MH & MH & MH & MH & MH & MH & \multirow{2}{*}{Avg} \\
& 000 & 001 & 002 & 003 & 004 & 005 & 006 & 007 & 000 & 001 & 002 & 003 & 004 & 005 & 006 & 007 & \\
\toprule
ORB-SLAM3*~\cite{campos2021orb} & 13.61 & 16.86 & 20.57 & 16.00 & 22.27 & 9.28 & 21.61 & 7.74 & 15.44 & 2.92 & 13.51 & 8.18 & 2.59 & 21.91 & 11.70 & 25.88 & 14.38 \\
COLMAP*~\cite{schonberger2016structure} & 15.20 & 5.58 & 10.86 & 3.93 & 2.62 & 14.78 & 7.00 & 18.47 & 12.26 & 13.45 & 13.45 & 20.95 & 24.97 & 16.79 & 7.01 & 7.97 & 12.50 \\
DSO~\cite{engel2017direct} & 9.65 & 3.84 & 12.20 & 8.17 & 9.27 & 2.94 & 8.15 & 5.43 & 9.92 & 0.35 & 7.96 & 3.46 & - & 12.58 & 8.42 & 7.50 & 7.32 \\
DROID-SLAM*~\cite{teed2021droid} & 0.17 & 0.06 & 0.36 & 0.87 & 1.14 & 0.13 & 1.13 & \textbf{0.06} & \textbf{0.08} & 0.05 & 0.04 & \textbf{0.02} & \textbf{0.01} & 0.68 & 0.30 & 0.07 & 0.33 \\
DROID-VO & 0.22 & 0.15 & 0.24 & 1.27 & 1.04 & 0.14 & 1.32 & 0.77 & 0.32 & 0.13 & 0.08 & 0.09 & 1.52 & 0.69 & 0.39 & 0.97 & 0.58 \\
\midrule
DPVO & 0.16 & 0.11 & 0.11 & 0.66 & 0.31 & 0.14 & 0.30 & 0.13 & 0.21 & 0.04 & 0.04 & 0.08 & 0.58 & \textbf{0.17} & 0.11 & 0.15 & 0.21 \\
Ours & \textbf{0.08} & \textbf{0.05} & 0.16 & \textbf{0.30} & \textbf{0.27} & \textbf{0.08} & \textbf{0.20} & 0.10 & 0.18 & \textbf{0.03} & \textbf{0.03} &\textbf{ 0.02} & 0.58 & 0.30 & \textbf{0.08} & \textbf{0.05} & \textbf{0.16} \\
\bottomrule
\end{tabular}
}
\caption{ATE [m] results on the TartanAir~\cite{wang2020tartanair} test split compared to other SLAM methods. For our method and DPVO, we report the median of 5 runs. (*) indicates the method used global loop closure optimization.\vspace{-1mm}}
\label{table:tartantest}
\end{table*}

\begin{table*}[h!]
\centering
\resizebox{0.8\textwidth}{!}{%
\begin{tabular}{l ccccc ccc ccc |c}
\toprule
& MH01 & MH02 & MH03 & MH04 & MH05 & V101 & V102 & V103 & V201 & V202 & V203 & Avg \\
\toprule
DPVO & 0.087 & 0.055 & \textbf{0.158} & \textbf{0.137} & 0.114 & 0.050 & 0.140 & \textbf{0.086} & 0.057 & \textbf{0.049} & 0.211 & \textbf{0.105} \\
Ours & \textbf{0.081} & 0.067 & 0.171 & 0.179 & 0.115 & \textbf{0.046} & 0.160 & 0.097 & 0.056 & 0.059 & 0.252 & 0.117 \\
\bottomrule
\end{tabular}
}
\caption{ATE [m] results on the EuRoC dataset \cite{burri2016euroc} compared to other visual odometry methods. For our method and DPVO, we report the median of 5 runs. The performance of our model is similar to DPVO.\vspace{-1mm}}
\label{table:euroc}
\end{table*}

\begin{table*}[h!]
\centering
\resizebox{0.65\textwidth}{!}{
\begin{tabular}{lccccccccc|c}
\toprule
& 360 & desk & desk2 & floor & plant & room & rpy & teddy & xyz & Avg \\
\midrule
DPVO & \textbf{0.135} & 0.038 & 0.048 & 0.040 & 0.036 & 0.394 & 0.034 & 0.064 & 0.012 & \textbf{0.089}\\
Ours & 0.145 & 0.026 & \textbf{0.044} & 0.064 & 0.031 & 0.434 & 0.045 & \textbf{0.046} & 0.012 & 0.094 \\
\bottomrule
\end{tabular}
}
\caption{ATE [m] results on the freiburg1 set of TUM-RGBD~\cite{sturm2012benchmark}. We evaluate \textit{monocular} visual odometry, and is identical to the evaluation setting in DPVO~\cite{teed2022deep}. For all methods, we report the median of 5 runs. (x) indicates that the method failed to track. The performance of our model is similar to DPVO.}
\label{thetable:tum_rgbd_results}
\end{table*}






\subsection{Test results for pose estimation}
We report pose estimation results on the TartanAir~\cite{wang2020tartanair} test-split from the CVPR 2020 SLAM competition in ~\autoref{table:tartantest}, and compare to results from other baseline methods as reported in DPVO~\cite{teed2022deep}. Traditional optimization-based approaches such as ORB-SLAM3~\cite{campos2021orb}, COLMAP~\cite{schonberger2016structure}, DSO~\cite{engel2017direct} fail to track accurately and have absolute trajectory errors (ATE) in the order of meters. Iterative learning-based DROID-SLAM~\cite{teed2021droid} and its variant without global loop-closure correction (DROID-VO) show reasonable performance, but DPVO is able to show much better accuracy by only tracking a sparse number of patches instead of dense flow. Our modified version, is able to show even better accuracy with a $~24\%$ lower error on average. Morever, we observe that our model outperforms DPVO on all but two sequences in the dataset. Using the same models trained on the TartanAir train set, we also report the results on the EuRoC~\cite{burri2016euroc} and the TUM-RGBD~\cite{sturm2012benchmark} benchmark datasets in ~\autoref{table:euroc} and ~\autoref{thetable:tum_rgbd_results}. Here, we obtain similar performance to DPVO. This suggests that, although the weighted flow loss helps improve the model accuracy on similar datasets, it doesn't resolve generalization issues related to domain shift from the TartanAir dataset to the real world. 



\section{Conclusions and Future work}

In this paper, we analyze the high variance in gradients during the training of pose estimation pipelines that use differentiable bundle adjustment layers. We identify three plausible causes for the high variance and show how they lead to slower training and instability. We then propose a simple solution for these problems 
involving a weighted correspondence loss. We implement this on a SOTA VO pipeline and demonstrate improved training stability and a $2.5$x training speedup. We also observe a $24\%$ accuracy improvement on the TartanAir test split and similar accuracy as the vanilla model on other benchmarks. Unsurprisingly, the modifications don't automatically improve the model's ability to tackle distribution shift. We also observe that the depth accuracy for low-weight points, which might be important for dense SLAM approaches, deteriorates. 

We see our work as an initial attempt at understanding the numerical issues stemming from the usage of bundle adjustment layers and optimization layers more broadly within deep learning pipelines. There are likely more factors contributing to issues like slower training, instability and generalization. We believe this broader area is relatively under-studied and requires more research to fully leverage the structure found in various real world problems.
{\small
\bibliographystyle{ieeenat_fullname}
\bibliography{egbib}

\begin{thebibliography}{61}
\providecommand{\natexlab}[1]{#1}
\providecommand{\url}[1]{\texttt{#1}}
\expandafter\ifx\csname urlstyle\endcsname\relax
  \providecommand{\doi}[1]{doi: #1}\else
  \providecommand{\doi}{doi: \begingroup \urlstyle{rm}\Url}\fi

\bibitem[Adler and {\"O}ktem(2017)]{adler2017solving}
Jonas Adler and Ozan {\"O}ktem.
\newblock Solving ill-posed inverse problems using iterative deep neural networks.
\newblock \emph{Inverse Problems}, 2017.

\bibitem[Adler and {\"O}ktem(2018)]{adler2018learned}
Jonas Adler and Ozan {\"O}ktem.
\newblock Learned primal-dual reconstruction.
\newblock \emph{IEEE transactions on medical imaging}, 2018.

\bibitem[Agrawal et~al.(2019)Agrawal, Amos, Barratt, Boyd, Diamond, and Kolter]{agrawal2019differentiable}
Akshay Agrawal, Brandon Amos, Shane Barratt, Stephen Boyd, Steven Diamond, and J~Zico Kolter.
\newblock Differentiable convex optimization layers.
\newblock \emph{Advances in neural information processing systems}, 32, 2019.

\bibitem[Amos and Kolter(2017)]{amos2017optnet}
Brandon Amos and J~Zico Kolter.
\newblock Optnet: Differentiable optimization as a layer in neural networks.
\newblock In \emph{International Conference on Machine Learning}, pages 136--145. PMLR, 2017.

\bibitem[Amos et~al.(2018)Amos, Jimenez, Sacks, Boots, and Kolter]{amos2018differentiable}
Brandon Amos, Ivan Jimenez, Jacob Sacks, Byron Boots, and J~Zico Kolter.
\newblock Differentiable mpc for end-to-end planning and control.
\newblock \emph{Advances in neural information processing systems}, 31, 2018.

\bibitem[Antonova et~al.(2023)Antonova, Yang, Jatavallabhula, and Bohg]{antonova2023rethinking}
Rika Antonova, Jingyun Yang, Krishna~Murthy Jatavallabhula, and Jeannette Bohg.
\newblock Rethinking optimization with differentiable simulation from a global perspective.
\newblock In \emph{Conference on Robot Learning}, pages 276--286. PMLR, 2023.

\bibitem[Barron(2019)]{Barron_2019_CVPR}
Jonathan~T. Barron.
\newblock A general and adaptive robust loss function.
\newblock In \emph{CVPR}, 2019.

\bibitem[Beck and Schmidt(2021)]{beck2021gentle}
Yasmine Beck and Martin Schmidt.
\newblock A gentle and incomplete introduction to bilevel optimization.
\newblock 2021.

\bibitem[Bhardwaj et~al.(2020)Bhardwaj, Boots, and Mukadam]{bhardwaj2020differentiable}
Mohak Bhardwaj, Byron Boots, and Mustafa Mukadam.
\newblock Differentiable gaussian process motion planning.
\newblock In \emph{2020 IEEE international conference on robotics and automation (ICRA)}, 2020.

\bibitem[Bianchini et~al.(2023)Bianchini, Halm, and Posa]{bianchini2023simultaneous}
Bibit Bianchini, Mathew Halm, and Michael Posa.
\newblock Simultaneous learning of contact and continuous dynamics.
\newblock \emph{arXiv preprint arXiv:2310.12054}, 2023.

\bibitem[Burnett et~al.(2021)Burnett, Yoon, Schoellig, and Barfoot]{burnett_rss21}
Keenan Burnett, David~J Yoon, Angela~P Schoellig, and Timothy~D Barfoot.
\newblock Radar odometry combining probabilistic estimation and unsupervised feature learning.
\newblock In \emph{Robotics: Science and Systems}, 2021.

\bibitem[Burri et~al.(2016)Burri, Nikolic, Gohl, Schneider, Rehder, Omari, Achtelik, and Siegwart]{burri2016euroc}
Michael Burri, Janosch Nikolic, Pascal Gohl, Thomas Schneider, Joern Rehder, Sammy Omari, Markus~W Achtelik, and Roland Siegwart.
\newblock The euroc micro aerial vehicle datasets.
\newblock \emph{The International Journal of Robotics Research}, 2016.

\bibitem[Cadena et~al.(2016)Cadena, Carlone, Carrillo, Latif, Scaramuzza, Neira, Reid, and Leonard]{cadena2016past}
Cesar Cadena, Luca Carlone, Henry Carrillo, Yasir Latif, Davide Scaramuzza, Jos{\'e} Neira, Ian Reid, and John~J Leonard.
\newblock Past, present, and future of simultaneous localization and mapping: Toward the robust-perception age.
\newblock \emph{IEEE Transactions on robotics}, 2016.

\bibitem[Campos et~al.(2021)Campos, Elvira, Rodr{\'\i}guez, Montiel, and Tard{\'o}s]{campos2021orb}
Carlos Campos, Richard Elvira, Juan J~G{\'o}mez Rodr{\'\i}guez, Jos{\'e}~MM Montiel, and Juan~D Tard{\'o}s.
\newblock Orb-slam3: An accurate open-source library for visual, visual--inertial, and multimap slam.
\newblock \emph{IEEE Transactions on Robotics}, 2021.

\bibitem[Chebrolu et~al.(2021)Chebrolu, L{\"a}be, Vysotska, Behley, and Stachniss]{chebrolu2021adaptive}
Nived Chebrolu, Thomas L{\"a}be, Olga Vysotska, Jens Behley, and Cyrill Stachniss.
\newblock Adaptive robust kernels for non-linear least squares problems.
\newblock \emph{IEEE Robotics and Automation Letters}, 2021.

\bibitem[Chen et~al.(2020{\natexlab{a}})Chen, Parra, Cao, Li, and Chin]{bpnp}
Bo Chen, Alvaro Parra, Jiewei Cao, Nan Li, and Tat-Jun Chin.
\newblock End-to-end learnable geometric vision by backpropagating pnp optimization.
\newblock In \emph{CVPR}, 2020{\natexlab{a}}.

\bibitem[Chen et~al.(2020{\natexlab{b}})Chen, Wang, Lu, Trigoni, and Markham]{chen2020survey}
Changhao Chen, Bing Wang, Chris~Xiaoxuan Lu, Niki Trigoni, and Andrew Markham.
\newblock A survey on deep learning for localization and mapping: Towards the age of spatial machine intelligence, 2020{\natexlab{b}}.

\bibitem[Clark et~al.(2018)Clark, Bloesch, Czarnowski, Leutenegger, and Davison]{clark2018learning}
Ronald Clark, Michael Bloesch, Jan Czarnowski, Stefan Leutenegger, and Andrew~J Davison.
\newblock Learning to solve nonlinear least squares for monocular stereo.
\newblock In \emph{ECCV}, 2018.

\bibitem[Czarnowski et~al.(2020)Czarnowski, Laidlow, Clark, and Davison]{deepfactors}
Jan Czarnowski, Tristan Laidlow, Ronald Clark, and Andrew~J Davison.
\newblock Deepfactors: Real-time probabilistic dense monocular slam.
\newblock \emph{IEEE Robotics and Automation Letters}, 2020.

\bibitem[DeTone et~al.(2018)DeTone, Malisiewicz, and Rabinovich]{detone2018superpoint}
Daniel DeTone, Tomasz Malisiewicz, and Andrew Rabinovich.
\newblock Superpoint: Self-supervised interest point detection and description.
\newblock In \emph{CVPR Workshops}, 2018.

\bibitem[Dontchev et~al.(2009)Dontchev, Rockafellar, and Rockafellar]{dontchev2009implicit}
Asen~L Dontchev, R~Tyrrell Rockafellar, and R~Tyrrell Rockafellar.
\newblock \emph{Implicit functions and solution mappings: A view from variational analysis}.
\newblock Springer, 2009.

\bibitem[Donti et~al.(2021)Donti, Rolnick, and Kolter]{donti2021dc}
Priya~L. Donti, David Rolnick, and J~Zico Kolter.
\newblock {DC}3: A learning method for optimization with hard constraints.
\newblock In \emph{International Conference on Learning Representations}, 2021.

\bibitem[Engel et~al.(2014)Engel, Sch{\"o}ps, and Cremers]{engel2014lsd}
Jakob Engel, Thomas Sch{\"o}ps, and Daniel Cremers.
\newblock Lsd-slam: Large-scale direct monocular slam.
\newblock In \emph{ECCV}, 2014.

\bibitem[Engel et~al.(2017)Engel, Koltun, and Cremers]{engel2017direct}
Jakob Engel, Vladlen Koltun, and Daniel Cremers.
\newblock Direct sparse odometry.
\newblock \emph{IEEE transactions on pattern analysis and machine intelligence}, 2017.

\bibitem[Fan et~al.(2022)Fan, Zhu, He, Sun, Liu, and He]{fan2022deep}
Zhaoxin Fan, Yazhi Zhu, Yulin He, Qi Sun, Hongyan Liu, and Jun He.
\newblock Deep learning on monocular object pose detection and tracking: A comprehensive overview.
\newblock \emph{ACM Computing Surveys}, 2022.

\bibitem[Fu and Fallon(2023)]{fu2023batch}
Lanke Frank~Tarimo Fu and Maurice Fallon.
\newblock Batch differentiable pose refinement for in-the-wild camera/lidar extrinsic calibration.
\newblock In \emph{Conference on Robot Learning}, 2023.

\bibitem[Fung et~al.(2022)Fung, Heaton, Li, McKenzie, Osher, and Yin]{fung2022jfb}
Samy~Wu Fung, Howard Heaton, Qiuwei Li, Daniel McKenzie, Stanley Osher, and Wotao Yin.
\newblock Jfb: Jacobian-free backpropagation for implicit networks.
\newblock In \emph{Proceedings of the AAAI Conference on Artificial Intelligence}, pages 6648--6656, 2022.

\bibitem[Howell et~al.(2022{\natexlab{a}})Howell, Le~Cleac’h, Kolter, Schwager, and Manchester]{howell2022dojo}
Taylor~A Howell, Simon Le~Cleac’h, J~Zico Kolter, Mac Schwager, and Zachary Manchester.
\newblock Dojo: A differentiable simulator for robotics.
\newblock \emph{arXiv preprint arXiv:2203.00806}, 9, 2022{\natexlab{a}}.

\bibitem[Howell et~al.(2022{\natexlab{b}})Howell, Tracy, Le~Cleac’h, and Manchester]{howell2022calipso}
Taylor~A Howell, Kevin Tracy, Simon Le~Cleac’h, and Zachary Manchester.
\newblock Calipso: A differentiable solver for trajectory optimization with conic and complementarity constraints.
\newblock In \emph{The International Symposium of Robotics Research}, pages 504--521. Springer, 2022{\natexlab{b}}.

\bibitem[Iwase et~al.(2021)Iwase, Liu, Khirodkar, Yokota, and Kitani]{iwase2021repose}
Shun Iwase, Xingyu Liu, Rawal Khirodkar, Rio Yokota, and Kris~M Kitani.
\newblock Repose: Fast 6d object pose refinement via deep texture rendering.
\newblock In \emph{ICCV}, 2021.

\bibitem[Jatavallabhula et~al.(2020)Jatavallabhula, Iyer, and Paull]{jatavallabhula2020slam}
Krishna~Murthy Jatavallabhula, Ganesh Iyer, and Liam Paull.
\newblock $\delta$ slam: Dense slam meets automatic differentiation.
\newblock In \emph{2020 IEEE International Conference on Robotics and Automation (ICRA)}, 2020.

\bibitem[Kanazawa and Kanatani(2001)]{kanazawa}
Y. Kanazawa and K. Kanatani.
\newblock Do we really have to consider covariance matrices for image features?
\newblock In \emph{ICCV}, 2001.

\bibitem[Koestler et~al.(2022)Koestler, Yang, Zeller, and Cremers]{koestler2022tandem}
Lukas Koestler, Nan Yang, Niclas Zeller, and Daniel Cremers.
\newblock Tandem: Tracking and dense mapping in real-time using deep multi-view stereo.
\newblock In \emph{Conference on Robot Learning}, 2022.

\bibitem[Labb{\'e} et~al.(2020)Labb{\'e}, Carpentier, Aubry, and Sivic]{labbe2020cosypose}
Yann Labb{\'e}, Justin Carpentier, Mathieu Aubry, and Josef Sivic.
\newblock Cosypose: Consistent multi-view multi-object 6d pose estimation.
\newblock In \emph{ECCV}, 2020.

\bibitem[Lepetit et~al.(2009)Lepetit, Moreno-Noguer, and Fua]{lepetit2009ep}
Vincent Lepetit, Francesc Moreno-Noguer, and Pascal Fua.
\newblock Ep n p: An accurate o (n) solution to the p n p problem.
\newblock \emph{IJCV}, 2009.

\bibitem[Lipson et~al.(2022)Lipson, Teed, Goyal, and Deng]{lipson2022coupled}
Lahav Lipson, Zachary Teed, Ankit Goyal, and Jia Deng.
\newblock Coupled iterative refinement for 6d multi-object pose estimation.
\newblock In \emph{CVPR}, 2022.

\bibitem[Lowe(2004)]{lowe2004distinctive}
David~G Lowe.
\newblock Distinctive image features from scale-invariant keypoints.
\newblock \emph{International journal of computer vision}, 2004.

\bibitem[Metz et~al.(2021)Metz, Freeman, Schoenholz, and Kachman]{metz2021gradients}
Luke Metz, C~Daniel Freeman, Samuel~S Schoenholz, and Tal Kachman.
\newblock Gradients are not all you need.
\newblock \emph{arXiv preprint arXiv:2111.05803}, 2021.

\bibitem[Min et~al.(2020)Min, Yang, and Dunn]{min2020voldor}
Zhixiang Min, Yiding Yang, and Enrique Dunn.
\newblock Voldor: Visual odometry from log-logistic dense optical flow residuals.
\newblock In \emph{Proceedings of the IEEE/CVF Conference on Computer Vision and Pattern Recognition}, 2020.

\bibitem[Muhle et~al.(2023)Muhle, Koestler, Jatavallabhula, and Cremers]{muhle2023learning}
Dominik Muhle, Lukas Koestler, Krishna~Murthy Jatavallabhula, and Daniel Cremers.
\newblock Learning correspondence uncertainty via differentiable nonlinear least squares.
\newblock In \emph{CVPR}, 2023.

\bibitem[Mur-Artal et~al.(2015)Mur-Artal, Montiel, and Tardos]{mur2015orb}
Raul Mur-Artal, Jose Maria~Martinez Montiel, and Juan~D Tardos.
\newblock Orb-slam: a versatile and accurate monocular slam system.
\newblock \emph{IEEE transactions on robotics}, 2015.

\bibitem[Park et~al.(2019)Park, Patten, and Vincze]{pix2pose}
Kiru Park, Timothy Patten, and Markus Vincze.
\newblock Pix2pose: Pixel-wise coordinate regression of objects for 6d pose estimation.
\newblock In \emph{ICCV}, 2019.

\bibitem[Pineda et~al.(2022)Pineda, Fan, Monge, Venkataraman, Sodhi, Chen, Ortiz, DeTone, Wang, Anderson, Dong, Amos, and Mukadam]{pineda2022theseus}
Luis Pineda, Taosha Fan, Maurizio Monge, Shobha Venkataraman, Paloma Sodhi, Ricky~TQ Chen, Joseph Ortiz, Daniel DeTone, Austin Wang, Stuart Anderson, Jing Dong, Brandon Amos, and Mustafa Mukadam.
\newblock {Theseus: A Library for Differentiable Nonlinear Optimization}.
\newblock \emph{Advances in Neural Information Processing Systems}, 2022.

\bibitem[Rad and Lepetit(2017)]{bb8}
Mahdi Rad and Vincent Lepetit.
\newblock Bb8: A scalable, accurate, robust to partial occlusion method for predicting the 3d poses of challenging objects without using depth.
\newblock In \emph{ICCV}, 2017.

\bibitem[Ranftl and Koltun(2018)]{ranftl2018deep}
Ren{\'e} Ranftl and Vladlen Koltun.
\newblock Deep fundamental matrix estimation.
\newblock In \emph{ECCV}, 2018.

\bibitem[Rublee et~al.(2011)Rublee, Rabaud, Konolige, and Bradski]{rublee2011orb}
Ethan Rublee, Vincent Rabaud, Kurt Konolige, and Gary Bradski.
\newblock Orb: An efficient alternative to sift or surf.
\newblock In \emph{ICCV}, 2011.

\bibitem[Sarlin et~al.(2020)Sarlin, DeTone, Malisiewicz, and Rabinovich]{sarlin2020superglue}
Paul-Edouard Sarlin, Daniel DeTone, Tomasz Malisiewicz, and Andrew Rabinovich.
\newblock Superglue: Learning feature matching with graph neural networks.
\newblock In \emph{CVPR}, 2020.

\bibitem[Schonberger and Frahm(2016)]{schonberger2016structure}
Johannes~L Schonberger and Jan-Michael Frahm.
\newblock Structure-from-motion revisited.
\newblock In \emph{CVPR}, 2016.

\bibitem[Sturm et~al.(2012)Sturm, Engelhard, Endres, Burgard, and Cremers]{sturm2012benchmark}
J{\"u}rgen Sturm, Nikolas Engelhard, Felix Endres, Wolfram Burgard, and Daniel Cremers.
\newblock A benchmark for the evaluation of rgb-d slam systems.
\newblock In \emph{2012 IEEE/RSJ international conference on intelligent robots and systems}, 2012.

\bibitem[Suh et~al.(2022)Suh, Simchowitz, Zhang, and Tedrake]{suh2022differentiable}
Hyung~Ju Suh, Max Simchowitz, Kaiqing Zhang, and Russ Tedrake.
\newblock Do differentiable simulators give better policy gradients?
\newblock In \emph{International Conference on Machine Learning}, pages 20668--20696. PMLR, 2022.

\bibitem[Tang and Tan(2019)]{tang2018ba}
Chengzhou Tang and Ping Tan.
\newblock {BA}-net: Dense bundle adjustment networks.
\newblock In \emph{ICLR}, 2019.

\bibitem[Tateno et~al.(2017)Tateno, Tombari, Laina, and Navab]{tateno2017cnn}
Keisuke Tateno, Federico Tombari, Iro Laina, and Nassir Navab.
\newblock Cnn-slam: Real-time dense monocular slam with learned depth prediction.
\newblock In \emph{CVPR}, 2017.

\bibitem[Teed and Deng(2021)]{teed2021droid}
Zachary Teed and Jia Deng.
\newblock {DROID}-{SLAM}: Deep visual {SLAM} for monocular, stereo, and {RGB}-d cameras.
\newblock In \emph{Advances in Neural Information Processing Systems}, 2021.

\bibitem[Teed et~al.(2023)Teed, Lipson, and Deng]{teed2022deep}
Zachary Teed, Lahav Lipson, and Jia Deng.
\newblock Deep patch visual odometry.
\newblock \emph{Advances in Neural Information Processing Systems}, 2023.

\bibitem[Ummenhofer et~al.(2017)Ummenhofer, Zhou, Uhrig, Mayer, Ilg, Dosovitskiy, and Brox]{Ummenhofer_2017_CVPR}
Benjamin Ummenhofer, Huizhong Zhou, Jonas Uhrig, Nikolaus Mayer, Eddy Ilg, Alexey Dosovitskiy, and Thomas Brox.
\newblock Demon: Depth and motion network for learning monocular stereo.
\newblock In \emph{CVPR}, 2017.

\bibitem[Wang et~al.(2017)Wang, Clark, Wen, and Trigoni]{deepvo}
Sen Wang, Ronald Clark, Hongkai Wen, and Niki Trigoni.
\newblock Deepvo: Towards end-to-end visual odometry with deep recurrent convolutional neural networks.
\newblock In \emph{IEEE Int. Conf. Robotics and Automation}, 2017.

\bibitem[Wang et~al.(2020)Wang, Zhu, Wang, Hu, Qiu, Wang, Hu, Kapoor, and Scherer]{wang2020tartanair}
Wenshan Wang, Delong Zhu, Xiangwei Wang, Yaoyu Hu, Yuheng Qiu, Chen Wang, Yafei Hu, Ashish Kapoor, and Sebastian Scherer.
\newblock Tartanair: A dataset to push the limits of visual slam.
\newblock In \emph{2020 IEEE/RSJ International Conference on Intelligent Robots and Systems (IROS)}, 2020.

\bibitem[Wang et~al.(2021)Wang, Hu, and Scherer]{tartanvo}
Wenshan Wang, Yaoyu Hu, and Sebastian Scherer.
\newblock Tartanvo: A generalizable learning-based vo.
\newblock In \emph{Conference on Robot Learning}, 2021.

\bibitem[Yang et~al.(2020)Yang, Stumberg, Wang, and Cremers]{d3vo}
Nan Yang, Lukas~von Stumberg, Rui Wang, and Daniel Cremers.
\newblock D3vo: Deep depth, deep pose and deep uncertainty for monocular visual odometry.
\newblock In \emph{CVPR}, 2020.

\bibitem[Zhou et~al.(2018)Zhou, Ummenhofer, and Brox]{zhou2018deeptam}
Huizhong Zhou, Benjamin Ummenhofer, and Thomas Brox.
\newblock Deeptam: Deep tracking and mapping.
\newblock In \emph{ECCV}, 2018.

\bibitem[Zhou et~al.(2017)Zhou, Brown, Snavely, and Lowe]{sfmlearner}
Tinghui Zhou, Matthew Brown, Noah Snavely, and David~G Lowe.
\newblock Unsupervised learning of depth and ego-motion from video.
\newblock In \emph{CVPR}, 2017.

\end{thebibliography}
}

\newpage

\section{Appendix}

\subsection{Acknowledgements}
SG was supported by a grant from the Bosch Center for Artificial Intelligence. KR was partially supported by the ERC Advanced Grant SIMULACRON. We would like to thank Zhengyang Geng, Michael Kaess, Dan Mccgann, Akash Sharma and numerous others for various brainstorming sessions and feedback during the initial stages of the project. We would also like to thank the authors of DPVO ~\cite{teed2021droid} and DROID-SLAM \cite{teed2021droid} for open sourcing their code and making it easy to use.

\subsection{Results on DROID-SLAM} \label{sec:droid}

\begin{figure}[h]
    \centering
    \includegraphics[scale=0.6]{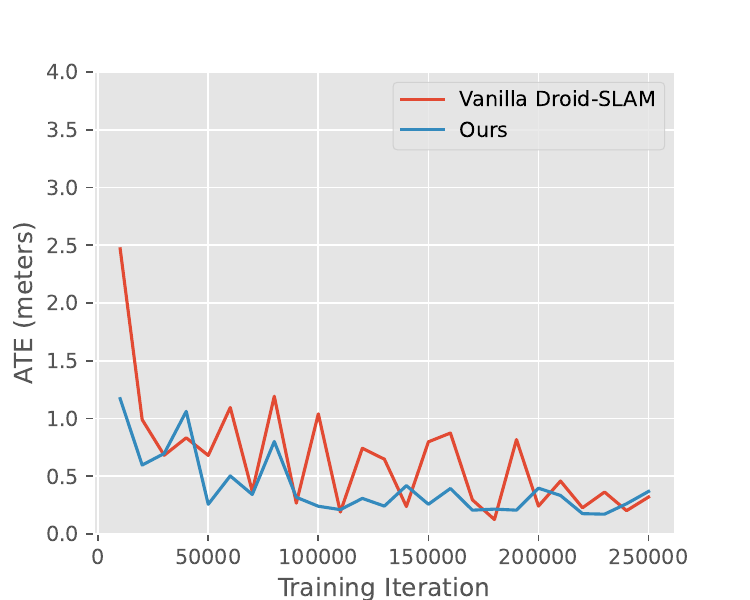}
    \caption{We plot the median validation ATE across three trials of DROID-SLAM and our modification at various iterations during training. We observe a clear speedup in training convergence and improved training stability suggesting that our method and analysis generalizes to other approaches using differentiable bundle adjustment layers.}
    \label{fig:droid-val}
\end{figure}

We test our modifications on DROID-SLAM. Specifically, we simply use a weighted flow loss \autoref{eq:wtdfloss} instead of their regular flow loss and tune the corresponding loss coefficient value. We keep the rest of the pipeline unchanged. \autoref{fig:droid-val} shows the validation ATEs throughout training for the baseline model and our modification. We clearly observe that our modification leads to a significant speedup as well as stabilization during training thanks to the reduced variance in the updates. This suggests that the methods and analysis discussed in this paper generalize broadly to a wide variety of approaches using differentiable bundle adjustment layers.

\subsection{Deriving the gradients for the BA problem} \label{sec:ift}

In this section, we derive the BA gradients given in \autoref{eq:flgrads} and \autoref{eq:wt_grads}. We start with the bundle adjustment objective given in \autoref{eqn:bundle_adjustment}
\begin{equation}
    \min_{\mathbf{T}_{ij}, d_k} \sum_{(k, j)} ||\bar{\mathbf{p}}_{jk}-\Pi(\mathbf{T}_{ij}, \Pi^{-1}(\mathbf{p}_k,d_{k}))||^2_{\Sigma_{jk}}
\end{equation}
For simplicity, we drop the subscripts and compute the gradients considering just one term in the above sum:
\begin{equation}
    \ell = \|\mathbf{r}(\mathbf{T}, d)\|^2_\Sigma 
\end{equation}
where $\mathbf{r}(\mathbf{T},d) = \bar{\mathbf{p}} - \hat{\mathbf{p}}(\mathbf{T},d)$, is the bundle adjustment residual, and $\hat{\mathbf{p}}(\mathbf{T},d) = \Pi(\mathbf{T}, \Pi^{-1}(\mathbf{p},d))$. We make a first order taylor series approximation of the residual $\mathbf{r}(\mathbf{T},d)$ around $\mathbf{T}, d$:
\begin{equation}
    \mathbf{r}(\mathbf{T},d) = \mathbf{r} + \mathbf{J}_{\mathbf{T}}\Delta \mathbf{T} + \mathbf{J}_d \Delta d
\end{equation}
 where, $\mathbf{J}_d$ and $\mathbf{J}_{\mathbf{T}}$ are the jacobians of the projection $\Pi(\mathbf{T}, \Pi^{-1}(\mathbf{p},d))$ with respect to corresponding depth $d$ and pose $\mathbf{T}$ respectively. 
Now, using the implicit function theorem, we can write the weight gradients as follows:
\begin{align}
    \nabla_\Sigma \mathcal{L} = &- (\nabla_\mathbf{T} \mathcal{L})^T \left(\frac{\partial^2 \ell}{\partial (\Delta \mathbf{T})^2}\right)^{-1} \frac{\partial^2 \ell}{\partial (\Delta \mathbf{T}) \partial \Sigma} \\
    &- (\nabla_d \mathcal{L})^T \left(\frac{\partial^2 \ell}{\partial (\Delta d)^2}\right)^{-1} \frac{\partial^2 \ell}{\partial (\Delta d) \partial \Sigma}
\end{align}

Substituting in the values for the partial derivatives using the taylor series approximation of the residual, we get:
\begin{align}
    \nabla_\Sigma \mathcal{L} &= -(\nabla_{\mathbf{T}}\mathcal{L})^T(\mathbf{J}_{\mathbf{T}}^T\Sigma \mathbf{J}_{\mathbf{T}})^{-1}\mathbf{J}_{\mathbf{T}}^T \text{diag}(\mathbf{r})\nonumber
    \\&
    \quad -(\nabla_{d}\mathcal{L})^T(\mathbf{J}_{d}^T\Sigma \mathbf{J}_{d})^{-1}\mathbf{J}_{d}^T\text{diag}(\mathbf{r}) 
\end{align}
where, the residual $\mathbf{r}$ is evaluated at the solved iterate $\hat{T}, \hat{d}$. 
Likewise, the gradient for the flow corrections $\delta$ are given by :
\begin{align}
    \nabla_\delta \mathcal{L} &= -(\nabla_{\mathbf{T}}\mathcal{L})^T(\mathbf{J}_{\mathbf{T}}^T\Sigma \mathbf{J}_{\mathbf{T}})^{-1}\mathbf{J}_{\mathbf{T}}^T\Sigma \nonumber
    \\&
    \quad -(\nabla_{d}\mathcal{L})^T(\mathbf{J}_{d}^T\Sigma \mathbf{J}_{d})^{-1}\mathbf{J}_{d}^T\Sigma
\end{align}

\subsection{Analyzing the effect of weight residual dependence} \label{sec:wt-res}

\begin{figure}
    \centering
    \includegraphics[scale=0.6]{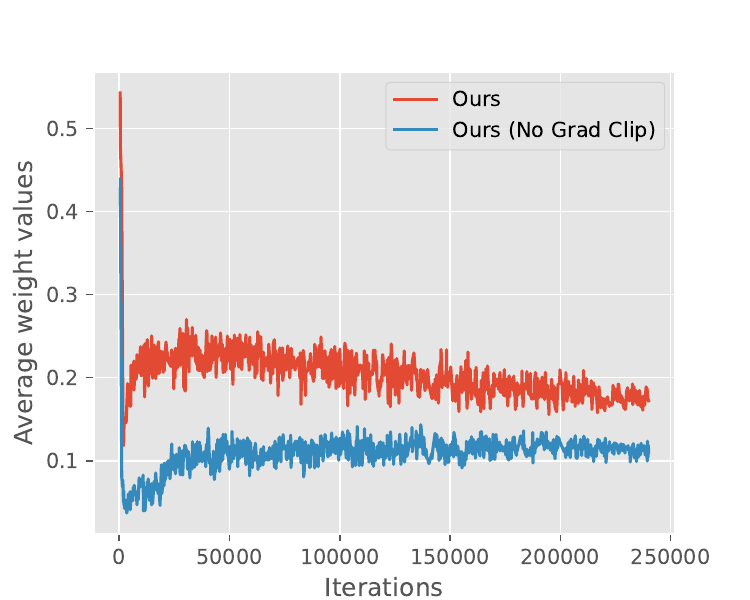}
    \caption{We plot the mean weight values of our method and our method without the weight gradient clipping throughout training. We observe a clear reduction in the weight magnitudes in the variant without gradient clipping suggesting a clear positive bias in the weight gradients in this variant. These ablations are performed on the non-streaming 8-step version of the problem.}
    \label{fig:wt-vals}
\end{figure}

\begin{figure}
    \centering
    \includegraphics[scale=0.6]{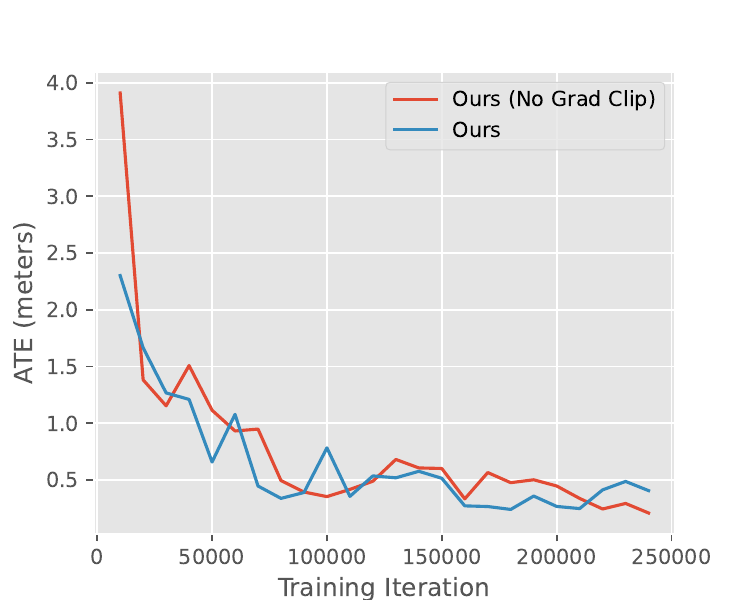}
    \caption{We plot the median validation ATE across three trials of our method and our method without the weight gradient clipping at various iterations during training. Both methods show similar performance suggesting that the architectures and inference method used in this paper is robust to the changes in weight distribution. These ablations are performed on the non-streaming 8-step version of the problem.}
    \label{fig:wt-valid}
\end{figure}

To understand the impact of the gradients on the weights, let's consider a simplified version of the weighted least squares problem as follows:
\begin{equation}
    f^* = \argmin_f \frac{1}{2}\sum_i \Sigma_i (f - \hat{f}_i)^2
\end{equation}
where $\Sigma_i$ and $f_i$ are the parameters of the optimization problem, with $\Sigma_i \in [0,1]$ being the weights. We further assume that we know the ground truth, $f_{\text{gt}}$, desired as the output of the optimization problem. We thus define the outer minimization problem as a minimization over the following loss:
\begin{equation} \label{eq:wt-grad}
    \min_{f_i, \Sigma_i \forall i} \mathcal{L} = \min_{f_i, \Sigma_i \forall i} | f_{\text{gt}} - f^* |
\end{equation}
The corresponding gradients of the outer loss $\mathcal{L}$ w.r.t the weights $\Sigma_i$ is given by :
\begin{equation}
    \nabla_{\Sigma_i} \mathcal{L} = - \text{sign}(f^* - f_{\text{gt}}) * \left(\frac{1}{\sum_i \Sigma_i}\right) * (f^* - \hat{f}_i).
\end{equation}
This gradient is positive when $(f^* - f_{\text{gt}})(f^* - \hat{f}_i) < 0$ given the weights are positive. In the presence of an outlier $f_k$ with high weight $\Sigma_k$, the least squares solution $f^*$ is biased towards $f_k$. Thus, the corresponding $(f^* - f_{\text{gt}})(f^* - \hat{f}_k) < 0$, thereby resulting in a highly positive gradient $\nabla_{\Sigma_k} \mathcal{L}$ given the high residual. 

In our original stochastic parameterized problem, each $\Sigma_i = \Sigma_\theta(x)$ is parameterized by a network followed by a sigmoid. Thus, the outliers with small $\Sigma_i$ don't contribute much to the gradient due to the sigmoid saturation. However, the outliers with large $\Sigma_i$ contribute significantly to the gradient. Given the outliers with large $\Sigma_i$ are inclined to get a highly positive gradient $\nabla_{\Sigma_k} \mathcal{L}$ as shown above, we observe an overall positive shift in the gradient values in the presence of outliers. If not handled carefully, this can potentially result in a collapse in the weight distribution to near zero values.

One of the architecture hacks used by \cite{teed2021droid, teed2022deep}, i.e, clipping the magnitude of gradient passing through the weights, incidentally already mitigates this issue to a large extent :
\begin{equation}
    \nabla_{w_i} \mathcal{L} := \text{max}(\text{min}(\nabla_{\Sigma_i} \mathcal{L}, \gamma_\text{max}), \gamma_\text{min})
\end{equation}
where $\gamma_\text{min}$ and $\gamma_\text{max}$ are the minimum and maximum gradient value thresholds. \autoref{fig:wt-vals} shows the average weight magnitudes for the last optimization iterate on two training runs, with and without the gradient clipping throughout training. We observe a clear reduction in the weight magnitudes as a result of removing the gradient clipping. This is especially pronounced in the initial iterations of training when the number of outliers are high. 

Interestingly, we observe that this change in the weight distribution does not have much effect on the performance of the model itself as shown in \autoref{fig:wt-valid}. The validation scores of the two models are similar throughout training. This phenomenon likely reflects the resilience inherent in the architecture and inference methodology used in this work. 

However, we believe that the positive bias in weight gradients can result in training instabilities in other settings that involve differentiable weighted least squares problems if not handled carefully. We thus include it in our discussions in this paper for a complete treatment although they don't seem to have a direct negative effect in the specific settings we investigate in this paper. 

\subsection{Weight collapse} \label{sec:wt-collapse}
The analysis in the previous section raises an interesting question about our specific setting, given that any collapse in weight values will significantly affect our training setup as the weights appear explicitly in the loss. However, as pointed earlier, empirically we don't observe any weight collapse. This raises two important questions. Why do the weights not collapse? Does our setting have any natural regularization that prevents such a weight collapse? We attempt to answer these questions as follows.

\begin{itemize}
    \item The positive bias in the weight gradients described in the previous section indeed exists resulting in the weight distribution being somewhat more conservative than necessary. However, this positive bias only results in this conservative offset instead of a total collapse in the weight values due to gradient clipping and the fraction of outliers being relatively minimal. Hence, this isn't problematic as a significant fraction of the weights still remain pretty high (\autoref{fig:wt-vals}) thereby providing sufficient training signal.
    \item We detach the weight values from the computation graph used in the loss to enforce the stop grad operation (Section 5). Hence, the loss grads don't directly contribute towards reduction of weight values. In fact, we show the analytical gradient w.r.t. the weights when differentiating through a simplified least squares problem in \autoref{eq:wt-grad}. An additional multiplication by the weight on these grads simply changes the individual grad magnitudes and doesn't change the direction. Furthermore, at a distributional level, we observe that the weight values between the two runs (weighted and unweighted) remain roughly similar throughout training.
    \item Due to the presence of both pose and flow loss, even if the weight values are reduced for some arbitrary reason, that would lead to a temporary reduction in the contribution of flow loss to the overall loss. This results in an increase in the contribution of pose loss gradients, acting as a buffer in case weights suddenly start collapsing (which we don’t observe empirically). However, we acknowledge that the first two \textit{inner-loop} iterations don't use pose loss and would not have this buffer. However, given that the procedure is iterative, the network would learn to correct for it from the third iteration onward.
\end{itemize}
\subsection{Effect of adding pose loss to the first few inner loop iterates} \label{sec:inner-pose}

\begin{figure}
    \centering
    \includegraphics[scale=0.6]{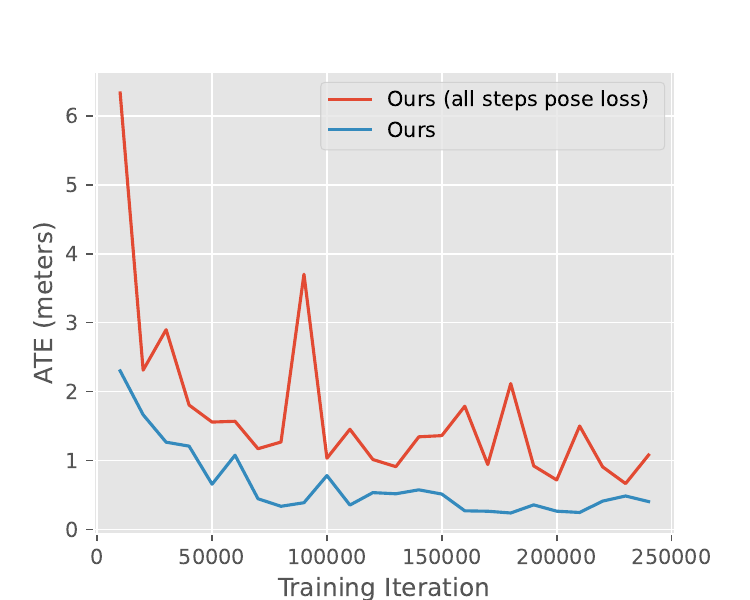}
    \caption{We plot the median validation ATE across three trials of our method and our method with pose loss added to all the iterates. We observe a clear deterioration in performance when adding the pose loss to the first two inner loop iterates as well. These ablations are performed on the non-streaming 8-step version of the problem.}
    \label{fig:pose-loss}
\end{figure}

As discussed in \autoref{sec:wtd-loss}, one of the tricks used by \cite{teed2021droid, teed2022deep}, already offer a partial solution to mask out the outliers in the loss, i.e., they simply do not add a pose loss on the first few inner loop iterates. Given the depth and pose estimates are especially bad in the first few inner loop iterations, this serves as an obvious solution to mitigate the gradient variance resulting from the linearization issues described in \autoref{sec:lingrads}. \autoref{fig:pose-loss} shows the effect of adding the pose loss on those initial iterates as well (in this case the first two iterates). We clearly observe a degradation in performance as a result due to the increased variance in gradients. 

\subsection{Alternative variance reduction methods} \label{sec:var-abl}

\begin{figure}
    \centering
    \includegraphics[scale=0.6]{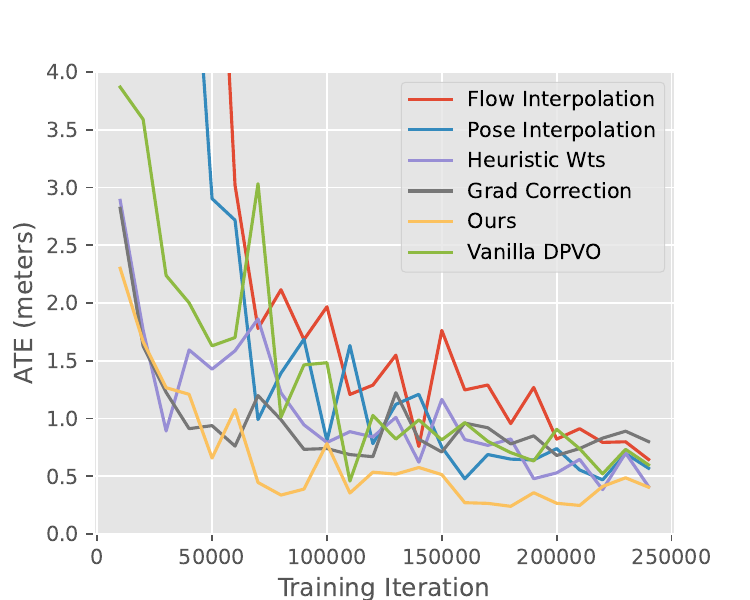}
    \caption{We plot the median validation ATE across three trials of alternate variance reduction strategies applied on the non-streaming 8-step version of DPVO.}
    \label{fig:var_red_abl}
\end{figure}

Although we propose one specific method for variance reduction in the paper (i.e using the weighted flow loss), we tested various other approaches towards variance reduction achieving varying degrees of success. We discuss a few of them in this section and compare them against our proposed modification. \autoref{fig:var_red_abl} shows the corresponding validation plots on the non-streaming 8-step version of the problem. 

\textit{\textbf{(a)} Removal of pose loss from initial inner-loop iterations}: As discussed in the previous section, removing the pose losses from the first few inner-loop iterations (\autoref{fig:pose-loss}) is very effective in reducing the variance and indeed we incorporate this change during our training.

\vspace{-2pt}
\textit{\textbf{(b)} Pose and flow interpolation using ground truth}: Given that we have access to ground truth flow and poses during training, we could interpolate between the pose/flow estimated by the network and the ground truth in order to reduce the variance of the iterates. We vary the interpolation coefficient such that we move from the ground truth distribution at the beginning of training to the model distribution towards the end of training. While this indeed reduces the variance, the interpolation adds a bias due to the distribution shift over training resulting from changing the interpolation coefficients. The other issue here arises from having to also interpolate the hidden state in an ad-hoc manner to account for the interpolated poses/flows.

\vspace{-2pt}
\textit{\textbf{(c)} Grad Correction using ground truth flows}: 
The ground truths can alternatively also be used to correct the gradients. Specifically, we clip the gradients on the flows at the input of the BA layer that aren't aligned with the ground truth ($\nabla_{\bar{\textbf{p}}_{jk}} \|\bar{\textbf{p}}_{jk} - \textbf{p}_{jk}^*\|$) to zero or reverse the sign. We observe that this and other similar methods for gradient correction bias the gradient, resulting in worse final performance, despite speeding up training in the initial iterations. One could potentially consider using this in the initial iterations of training to obtain the initial speedups and then switching to regular training.

\vspace{-2pt}
\textit{\textbf{(d)} (Heuristic) weights using ground truth}: 
We experimented with a number of heuristic weights that down-weight the outlier flows in the flow loss (\autoref{eq:wtdfloss}) based on their distance from the ground truth. We plot one such variant in \autoref{fig:var_red_abl} that uses the following heuristic for weight computation in the flow loss:
\begin{equation}
    \Sigma_{ijk} = \frac{1}{4 (p_{ijk}^* - \hat{p}_{ijk})/m + 1)}; \quad m = \text{median}(\mathbf{p}^* - \mathbf{\hat{p}})
\end{equation}
 We observe an improvement over the baseline, but couldn't find any heuristic that worked better than using the network-predicted weights (ours).

 The experiments in this section show that although there exist various methods for variance reduction, finding the right combination that does not add unintended bias to the gradients is a challenging task. This underscores the value of our analysis and the significance of the performance boost provided by our method.

\subsection{SNR computation} \label{sec:snr}
We use SNR values of the gradients in \autoref{fig:linearization} and \autoref{fig:snr-wtd} to analyze the noise levels in the gradients. In this section, we discuss the specific details of SNR computation in these two cases. In \autoref{fig:linearization}, we compute SNR (db) as follows:
\begin{equation}
    \text{SNR (db)} = 10 \log_{10} (\frac{\text{signal}}{\text{noise}})
\end{equation}
where signal$=\|\nabla_\delta\|$ is the magnitude of the gradient computed when no noise is added to the depth, and noise$=\|\hat{\nabla_\delta} - \nabla_\delta\|$ is the magnitude of the difference between the noisy and clean gradient.

In \autoref{fig:snr-wtd}, we report the SNR directly as :
\begin{align}
    \text{SNR} = (\frac{\text{signal}}{\text{noise}})
\end{align}
where signal$=\|\text{mean\_grad}\|$ is the magnitude of the mean gradient across a batch of examples, where $\text{mean\_grad}=\sum_{i=1}^N \frac{(\nabla_\theta)_i}{N}$. The noise$=\sum_{i=1}^N\frac{\|(\nabla_\theta)_i - \text{mean\_grad}\|}{N}$ is the average magnitude of the deviations from the mean gradient.
\end{document}